\documentclass[10pt,journal,compsoc]{IEEEtran}
\usepackage[utf8]{inputenc} 
\usepackage[T1]{fontenc}    
\usepackage{hyperref}       
\usepackage{url}            
\usepackage{booktabs}       
\usepackage{amsfonts}       
\usepackage{nicefrac}       
\usepackage{microtype}      
\usepackage{amsmath,bm}
\usepackage{graphicx}
\usepackage{amsthm}
\usepackage{algorithm}
\usepackage{algorithmic}
\usepackage{siunitx}
\usepackage{textcomp}
\usepackage{mathrsfs}
\usepackage{tikz}
\usepackage{textcomp}
\usepackage{subfigure}
\usepackage{placeins}
\newtheorem{theorem}{Theorem}
\usepackage[T1]{fontenc}
\usepackage{thmtools}
\usepackage{booktabs}
\usepackage{threeparttable}
\usepackage{hyperref}
\usepackage{multirow}
\usepackage{algorithm}
\usepackage{algorithmic}
\usepackage{xcolor,colortbl}
\definecolor{Gray}{gray}{0.9}
\newcommand{\grayline}{\raisebox{2pt}{\tikz{\draw[-,black!40!gray,solid,line width = 0.9pt](0,0) -- (6mm,0);}}}
\newcommand{\blueline}{\raisebox{2pt}{\tikz{\draw[-,blue,dashed,line width = 1.5pt](0,0) -- (6mm,0);}}}
\newcommand{\greenline}{\raisebox{2pt}{\tikz{\draw[-,green,dashed,line width = 1.5pt](0,0) -- (6mm,0);}}}
\newcommand{\redline}{\raisebox{2pt}{\tikz{\draw[-,red,dashed,line width = 1.5pt](0,0) -- (6mm,0);}}}
\newcommand{\blackline}{\raisebox{2pt}{\tikz{\draw[-, black, dashed,line width = 1.5pt](0,0) -- (6mm,0);}}}
\newcommand{\purplerectangle}{\begin{tikzpicture}
    \filldraw[fill=blue] (0,0) -- (0.1,0.1732) -- (0.2,0) -- cycle;
    \end{tikzpicture}}

\newcommand{\vect}[1]{\mathbf{#1}}

\newcommand{\eg}{e.g.\@}
\newcommand{\ie}{i.e.\@}
\newcommand{\etal}{et al.\@}

\newcommand{\mini}{\textit{mini}ImageNet}

\newcommand{\cifarfs}{\textsc{cifar-fs}}

\newcommand{\lstm}{\textsc{lstm}}

\usepackage{amsmath,amsfonts,bm}

\def\eqref#1{equation~\ref{#1}}

\def\1{\bm{1}}

\DeclareMathAlphabet{\mathsfit}{\encodingdefault}{\sfdefault}{m}{sl}
\SetMathAlphabet{\mathsfit}{bold}{\encodingdefault}{\sfdefault}{bx}{n}

\def\gX{{\mathcal{X}}}
\def\gY{{\mathcal{Y}}}
\def\gZ{{\mathcal{Z}}}

\newcommand{\E}{\mathbb{E}}

\newcommand{\softmax}{\mathrm{softmax}}

\newcommand{\KL}{D_{\mathrm{KL}}}

\DeclareMathOperator*{\argmin}{arg\,min}

\DeclareMathOperator{\Tr}{Tr}

\usepackage{ragged2e}
\usepackage{makecell} 
\usepackage{cleveref}
\crefformat{section}{\S#2#1#3} 
\crefformat{subsection}{\S#2#1#3}
\crefformat{subsubsection}{\S#2#1#3}
\usepackage{xcolor}
\usepackage{overpic}

\def\ie{\emph{i.e.}}
\def\eg{\emph{e.g.}}

\def\etal{{\em et al.}}

\usepackage{color}

\begin{document}

%
\title{MetaKernel: Learning Variational Random Features with Limited Labels}

%
%

%
%
%
%

\author{Yingjun~Du,
        Haoliang~Sun,
        Xiantong~Zhen,
        Jun~Xu,
        Yilong~Yin,
       Ling~Shao,~\IEEEmembership{Fellow,~IEEE}
       and~Cees~G.~M.~Snoek,~\IEEEmembership{Senior Member,~IEEE}
\IEEEcompsocitemizethanks{
\IEEEcompsocthanksitem Y. Du, X. Zhen and C. Snoek are with the University of Amsterdam, Amsterdam, the Netherlands.\protect
\IEEEcompsocthanksitem J. Xu is with Nankai University, Tianjin, China. \protect
\IEEEcompsocthanksitem H. Sun, and Y. Yin are with the School of Software, Shandong University, China. \protect
\IEEEcompsocthanksitem X. Zhen and L. Shao are with the Inception Institute of Artificial Intelligence, UAE.
\IEEEcompsocthanksitem 
A preliminary version of this work appeared in ICML 2020~\cite{zhen2020learning}.
}
\thanks{Manuscript received , 2021; revised , 2021.}}

%
%

\markboth{}%
{\Phihen \MakeLowercase{\textit{et al.}}: Bare Demo of IEEEtran.cls for Computer Society Journals}
%


\IEEEtitleabstractindextext{%
\justify{\begin{abstract}
Few-shot learning deals with the fundamental and challenging problem of learning from a few annotated samples, while being able to generalize well on new tasks. The crux of few-shot learning is to extract prior knowledge from related tasks to enable fast adaptation to a new task with a limited amount of data. In this paper, we propose meta-learning kernels with random Fourier features for few-shot learning, we call MetaKernel. Specifically, we propose learning variational random features in a data-driven manner to obtain task-specific kernels by leveraging the shared knowledge provided by related tasks in a meta-learning setting. We treat the random feature basis as the latent variable, which is estimated by variational inference. The shared knowledge from related tasks is incorporated into a context inference of the posterior, which we achieve via a long-short term memory module. To establish more expressive kernels, we deploy conditional normalizing flows based on coupling layers to achieve a richer posterior distribution over random Fourier bases. The resultant kernels are more informative and discriminative, which further improves the few-shot learning. To evaluate our method, we conduct extensive experiments on both few-shot image classification and regression tasks. A thorough ablation study demonstrates that the effectiveness of  each introduced component in our method. The benchmark results on fourteen datasets demonstrate MetaKernel consistently delivers at least comparable and often better performance than state-of-the-art alternatives. 
\end{abstract}}

}

\maketitle

\IEEEdisplaynontitleabstractindextext

\IEEEpeerreviewmaketitle

\IEEEraisesectionheading{\section{Introduction}\label{sec:introduction}}
\IEEEPARstart{H}{umans} have the amazing ability to learn new concepts from only a few examples, and then effortlessly generalize this knowledge  to new samples. In contrast, despite considerable progress, existing image classification models based on deep neural networks \eg,~\cite{krizhevsky2012imagenet,resnet}, are still highly dependent on large amounts of annotated training data~\cite{imagenet_cvpr09} to achieve satisfactory performance. This learnability gap between human intelligence and existing neural networks has motivated many to study learning from a few samples, \eg,~\cite{fei2006one,lake2015human,ravi2017optimization,finn2017model}. Meta-learning, \textit{a.k.a.} learning to learn~\cite{Schmidhuber1992,thrun2012learning}, emerged as a promising direction for few-shot learning~\cite{andrychowicz2016learning,ravi2017optimization,finn2017model,zhen2020learning}.

The working mechanism of meta-learning involves a meta-learner that exploits the common knowledge from various tasks to improve the performance of each individual task. Remarkable success has been achieved in learning good parameter initializations~\cite{finn2017model,rusu2018meta},  efficient optimization update rules~\cite{andrychowicz2016learning, ravi2017optimization}, and powerful common metrics~\cite{vinyals2016matching, snell2017prototypical} from related tasks, which enables fast adaptation to new tasks with few training samples. Meta-learning has also proven to be effective in learning amortized networks shared by related tasks, which generate specific parameters \cite{gordon2018meta} or normalization statistics \cite{du2020metanorm} for individual few-shot learning tasks. However, how to properly define and exploit the prior knowledge from experienced tasks remains an open problem for few-shot learning, and is the one we address in this paper.

An effective base-learner should be powerful enough to solve individual tasks, while being able to absorb the information provided by the meta-learner for overall benefit. Kernels \cite{smola1998learning,scholkopf2018learning,hofmann2008kernel} have proven to be a powerful technique in the machine learning toolbox, \eg,~\cite{cristianini2000introduction,smola2004tutorial,rahimi2007random,sinha2016learning,bach2004multiple}, as they are able to produce strong performance without relying on a large amount of labelled data. 
Moreover, task-adaptive kernels with random features, leveraging data-driven sampling strategies~\cite{sinha2016learning}, achieve improved performance over universal ones, at low sampling rates~\cite{hensman2017variational,carratino2018learning,bullins2018not,li2019implicit}. This makes kernels with data-driven random features well-suited tools for learning tasks with limited data. Hence, we introduce kernels as base-learners into the meta-learning framework for few-shot learning.
However, due to the limited availability of samples, it is challenging to learn informative random features for few-shot tasks by solely relying on a tasks own data. Therefore, exploring the shared prior knowledge from different but related tasks is essential for obtaining richer random features and few-shot learning.

\begin{figure*}[t]
	\centering
	\includegraphics[width=.9\linewidth]{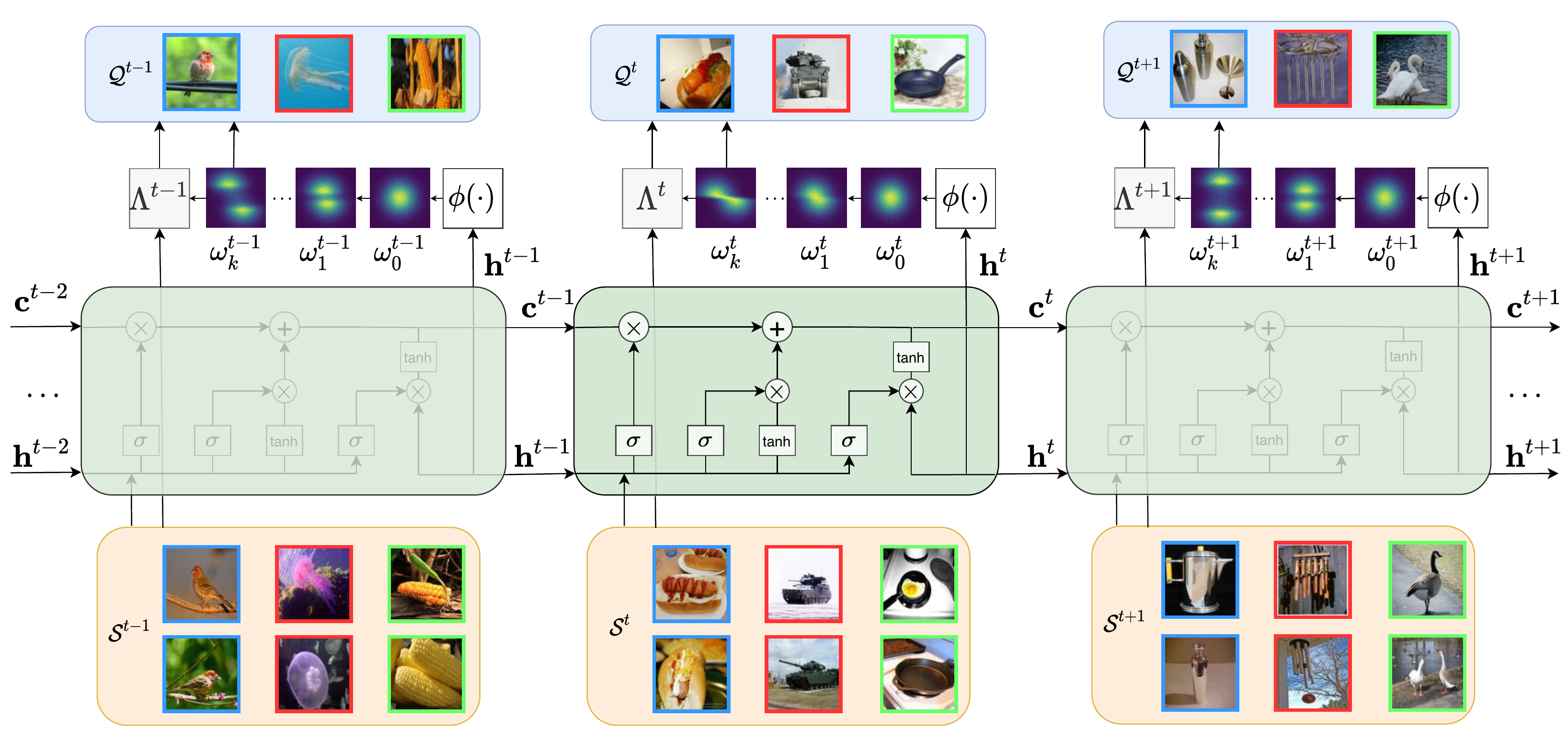}
	\vspace{-2mm}
	\caption{MetaKernel learning framework. The meta-learner employs an LSTM-based context inference network $\phi(\cdot)$ to infer the spectral distribution over $\bm{\omega}_0^{t}$, the kernel from the support set $\mathcal{S}^t$ of the current task $t$, and the outputs $\mathbf{h}^{t-1}$ and $\mathbf{c}^{t-1}$ of the previous task. The enriched random bases $\bm{\omega}_k^{t}$ are obtained via conditional normalizing flows with a flow of length $k$. During the learning process, the cell state in the LSTM is deployed to accumulate shared knowledge by experiencing a set of prior tasks. The \textit{remember} and \textit{forget} gates in the LSTM episodically refine the cell state by absorbing information from each experienced task. For each individual task, the task-specific information extracted from the support set is combined with distilled information from the previous tasks to infer the adaptive spectral distribution of the kernels.} 
	\label{fig:MeteKernel}
\end{figure*}

We propose learning task-specific kernels in a data-driven way with variational random features by leveraging the shared knowledge provided by related tasks. To do so, we develop a latent variable model that treats the random Fourier basis of translation-invariant kernels as the latent variable. The posterior over the random feature basis corresponds to the spectral distribution associated with the kernel. The optimization of the model is formulated as a variational inference problem. Kernel learning with random Fourier features for few-shot learning allows us to leverage the universal approximation property of kernels to capture shared knowledge from related tasks. This probabilistic modelling framework provides a principled way of learning data-driven kernels with random Fourier features and, more importantly, fits well into the meta-learning framework for few-shot learning, providing us the flexibility to customize the variational posterior and leverage meta-knowledge to enhance individual tasks.

To incorporate the prior knowledge from experienced tasks, we further propose a context inference scheme to integrate the inference of random feature bases of the current task into the context of previous related tasks. The context inference provides a generalized way to integrate shared knowledge from the related tasks with task-specific information for the inference of random feature bases. To do so, we adopt a long short-term memory (LSTM) based inference network~\cite{hochreiter1997long}, leveraging its capability of learning long-term dependencies to collect and refine the shared meta-knowledge from a set of previously experienced tasks.

A preliminary conference version of this work, which  also covers variational random features and task context inference was published previously~\cite{zhen2020learning}. In this extended work, we further propose conditional normalizing flows to infer richer posteriors over the random bases, which allows us to obtain more informative random features. The normalizing flows (NFs)~\cite{dinh2014nice, dinh2016density, rezende2015variational, kingma2018glow, winkler2019learning} model complicated high dimensional marginal distributions by transforming a simple base distribution (\eg, a standard normal) or priors through a learnable, invertible mapping and then applying the change of variables formula. Normalizing flows, which have not yet been explored in few-shot learning, provide a well-suited technique for learning more expressive random features by transforming a random basis into a richer distribution. The overall learning framework of our MetaKernel is illustrated in Figure~\ref{fig:MeteKernel}.

To validate our method, we conduct extensive experiments on fourteen benchmark datasets for a variety of few-shot learning tasks including image classification and regression. Unlike our prior work~\cite{zhen2020learning}, we also experiment on the large-scale Meta-Dataset by Triantafillou \etal~\cite{triantafillou2019meta} and the challenging few-shot domain generalization setting suggested by Du \etal~\cite{du2020metanorm}. MetaKernel consistently delivers at least comparable and often better performance than state-of-the-art alternatives on all datasets, and the ablative analysis demonstrates the effectiveness of each MetaKernel component for few-shot learning.

The rest of this paper is organized as follows:
Section~\ref{sec:related} summarizes related work. Section~\ref{sec:method} presents the proposed MetaKernel framework. Section~\ref{sec:experiments} summarizes experimental details, state-of-the-art comparisons and detailed ablation studies. Section~\ref{sec:conclusion} closes with concluding remarks.

\section{Related Work} 
\label{sec:related}

\subsection{Meta-Learning}
Meta-learning, or learning to learn, endues machine learning models the ability to improve their performance by leveraging knowledge extracted from a number of prior tasks. It has received increasing research interest with breakthroughs in many directions, \eg,~\cite{finn2017model,rusu2018meta,gordon2018meta,rajeswaran2019meta, hospedales2020meta}. Existing methods can be roughly categorized into four groups.

Models in the first group are based on distance metrics and generally learn a shared or adaptive embedding space in which query images are accurately matched to support images for classification. They rely on the assumption that a common metric space is shared across related tasks and usually do not employ an explicit base-learner for each task. By extending the matching network~\cite{vinyals2016matching} to few-shot scenarios, Snell \etal~\cite{snell2017prototypical} constructed a prototype for each class by averaging the feature representations of samples from the class in the metric space. The classification is conducted by matching the query samples to prototypes by computing their distances. To enhance the prototype representation, Allen \etal~\cite{allen2019infinite} proposed an infinite mixture of prototypes (IMP) to adaptively represent data distributions for each class, using multiple clusters instead of a single vector. Oreshkin \etal~\cite{oreshkin2018tadam} proposed a task-dependent adaptive metric for few-shot learning and established prototypes of classes conditioned on a task representation encoded by a task embedding network. Yoon \etal~\cite{yoon2019tapnet} proposed a few-shot learning algorithm aided by a linear transformer that performs task-specific null-space projection of the network output. Graphical neural network based models generalize the matching methods by learning the message propagation from the support set and transferring it to the query set~\cite{garcia2018few}. Prototype based methods have recently been improved in a variety of ways \cite{cao2019theoretical,triantafillou2019meta,zhen2020memory}.
In this work, we design an explicit base-learner based on kernels for each individual task.

Algorithms in the second group learn an optimization that is shared across tasks, while being adaptable to new tasks. Finn \etal~\cite{finn2017model} proposed model-agnostic meta-learning (MAML) to learn an appropriate initialization of model parameters and adapt it to new tasks with only a few gradient steps. To make MAML less prone to meta-overfitting, easier to parallelize and more interpretable, Zintgraf \etal~\cite{zintgraf2019fast} proposed fast context adaptation via meta-learning (CAVIA), a single model that adapts to a new task via gradient descent by updating only a set of input parameters at test time, instead of the entire network. Ravi and Larochelle~\cite{ravi2017optimization}  proposed an LSTM-based meta-learner that is trained to optimize a neural network classifier. It captures both the short-term knowledge in individual tasks and the long-term knowledge common to all tasks. Learning a shared optimization algorithm has also been explored to quickly learn new tasks~\cite{andrychowicz2016learning,chen2017learning}. Bayesian meta-learning methods~\cite{edwards2016towards,finn2018probabilistic, gordon2018meta,saemundsson2018meta} usually rely on hierarchical Bayesian models to learn the shared statistical information from different tasks and to infer the uncertainty of the models. Rusu \etal~\cite{rusu2018meta} proposed to learn a low-dimensional latent embedding of model parameters and perform optimization-based meta-learning in this space, which allows for a task-specific parameter initialization and achieves adaptation more effectively. Our method is orthogonal to optimization based methods and learns a specific base-learner for each task.

The third group of explicitly learned base-learners incorporate what meta-learners have learned and effectively addresses individual tasks~\cite{gordon2018meta, bertinetto2018meta, zhen2020learning}. Gordon \etal~\cite{gordon2018meta} avoided the need for gradient based optimization at test time by amortizing the posterior inference of task-specific parameters in their VERSA. It amortizes the cost of inference and alleviates the need for second derivatives during training by replacing test-time optimization with a forward pass through the inference network.  To enable efficient adaptation to unseen learning problems, Bertinetto \etal~\cite{bertinetto2018meta} incorporated fast solvers with closed-form solutions as the base learning component of their meta-learning framework. These teach the deep network to use ridge regression as part of its own internal model, enabling it to quickly adapt to novel data. In our method, we also deploy an explicit base-learner but, differently, we leverage a memory mechanism based on an LSTM to collect shared knowledge from related tasks and enhance the base-learners for individual tasks.

In the fourth group, a memory mechanism is part of the solution, where an external memory module is deployed to store and leverage key knowledge for quick adaptation~\cite{santoro2016meta,munkhdalai2017meta, munkhdalai2017rapid}. Santoro \etal~\cite{santoro2016meta} introduced neural Turing machines into meta-learning by augmenting their neural network with an external memory module, which is used to rapidly assimilate new data to  help make accurate predictions with only a few samples. Munkhdalai \etal~\cite{munkhdalai2017meta} proposed a Meta Network (MetaNet) to learn meta-level knowledge across tasks and shifting the inductive biases via fast parameterization for rapid generalization. Munkhdalai \etal~\cite{munkhdalai2017rapid} designed conditionally shifted neurons within the framework of meta-learning, which modify their activation values with task-specific shifts retrieved from a memory module. In this work, we also leverage a memory mechanism, but, differently, we deploy an LSTM module to collect shared knowledge from related tasks experienced previously to help solve individual tasks.


\subsection{Kernel Learning}
Kernel learning with random Fourier features is a versatile and powerful tool in machine learning~\cite{bishop2006pattern, hofmann2008kernel, shervashidze2011weisfeiler}. Pioneering works~\cite{bach2004multiple,gonen2011multiple, duvenaud2013structure} learn to combine predefined kernels in a multi-kernel learning manner. Kernel approximation by random Fourier features (RFFs)~\cite{rahimi2008random} is an effective technique for efficient kernel learning~\cite{gartner2002multi}, which has recently become increasingly popular~\cite{sinha2016learning,carratino2018learning}. RFFs~\cite{rahimi2008random} are derived from Bochner's theorem~\cite{rudin1962fourier}.
\begin{theorem}[Bochner's theorem~\cite{rudin1962fourier}] A continuous, real valued, symmetric and shift-invariant function $\mathtt{k}(\mathbf{x},\mathbf{x}') = \mathtt{k}(\mathbf{x}-\mathbf{x}')$ on $\mathbb{R}^d$ is a positive definite kernel if and only if it is the Fourier transform $p(\bm{\omega})$ of a positive finite measure such that
\begin{align}
\mathtt{k}(\mathbf{x},\mathbf{x}') =& \int_{\mathbb{R}^d} e^{i\bm{\omega}^\top(\mathbf{x}-\mathbf{x}')}dp(\bm{\omega}) = \mathbb{E}_{\bm{\omega}}[\zeta_{\bm{\omega}}(\mathbf{x})\zeta_{\bm{\omega}}(\mathbf{x}')^*]
\end{align}
where $\zeta_{\bm{\omega}}(\mathbf{x}) = e^{i\bm{\omega}^\top \mathbf{x}}$.
\end{theorem}
It is guaranteed that $\zeta_{\bm{\omega}}(\mathbf{x})\zeta_{\bm{\omega}}(\mathbf{x})^*$ is an unbiased estimation of $\mathtt{k}(\mathbf{x}, \mathbf{x}')$ with sufficient RFF bases $\{\bm{\omega}\}$ drawn from $p(\bm{\omega})$~\cite{rahimi2008random}.
For a predefined kernel, \eg, radial basis function (RBF), we sample from its spectral distribution using the Monte Carlo method, and obtain the explicit feature map:
\begin{equation}
\mathbf{z}(\mathbf{x}) = \frac{1}{\sqrt{D}} [\cos(\bm{\omega}_1^{\top} \mathbf{x} + b_1), \cdots, \cos(\bm{\omega}_D^{\top} \mathbf{x} + b_D)],
\label{rfs}
\end{equation}
where $\{\bm{\omega}_1, \cdots, \bm{\omega}_D\}$ are the random bases sampled from $p(\bm{\omega})$, and $[b_1, \cdots, b_D]$ are $D$ biases sampled from a uniform distribution with a range of $[0, 2\pi]$. 
Finally, the kernel value $\mathtt{k}(\mathbf{x}, \mathbf{x}')=\mathbf{z}(\mathbf{x})\mathbf{z}(\mathbf{x}')^{\top}$ in $K$ is computed as the dot product of their random feature maps with the same bases. 

Wilson and Adams~\cite{wilson2013gaussian} learn kernels in the frequency domain by modelling the spectral distribution as a mixture of Gaussians and computingits optimal linear combination. Instead of modelling the spectral distribution with explicit density functions, other works focus on optimizing the random base sampling strategy~\cite{yang2015carte, sinha2016learning}. Nonetheless, it has been shown that accurate approximation of kernels does not necessarily result in high classification performance \cite{avron2016quasi,chang2017data}. This suggests that learning adaptive kernels with random features by data-driven sampling strategies \cite{sinha2016learning} can improve the performance, even with a low sampling rate, compared to using universal random features \cite{avron2016quasi,chang2017data}.

Our work introduces kernels into few-shot meta-learning. We propose to learn kernels with random features in a data-driven way by formulating it as a variational inference problem. This allows us to generate task-specific kernels as well as to leverage shared knowledge from related tasks.

\subsection{Normalizing Flows}


Normalizing flows (NFs)~\cite{papamakarios2021normalizing,dinh2014nice,rezende2015variational} are promising methods for expressive probability density estimation with tractable distributions. Unlike variational methods, sampling and density evaluation can be efficient and exact for NFs with neat architectures. Generally, NFs are categorized into five types based on how they construct a flow: 1) Autoregressive flows were one of the first classes of flows with invertible autoregressive functions. Examples of such flows include inverse autoregressive flow~\cite{kingma2016improved} and masked autoregressive flow~\cite{papamakarios2017masked}. 2) Linear flows generalize the idea of permutating of input variables via an invertible linear transformation~\cite{kingma2018glow}. 3) Residual flows~\cite{chen2019residual} are designed as residual networks. The invertible property can be preserved under appropriate constraints; 4) volume-preserved flows with effective invertible architecture, such as coupling layers~\cite{dinh2016density}, are typically used in generative tasks. 5) Infinitesimal flows provide another alternative strategy for constructing flows in continuous time by parameterizing its infinitesimal dynamics~\cite{rezende2015variational}. Normalizing flows are known to be effective in applications with probabilistic models, 
including probabilistic modelling~\cite{kingma2018glow, ho2019flow++,esling2019universal, prenger2019waveglow}, inference \cite{rezende2015variational,kingma2016improved} and representation learning~\cite{jacobsen2018revnet}. 

In this work, we introduce conditional normalizing flows into our kernel learning framework to infer richer posteriors over the random bases, which yields  more informative random features. To our knowledge, this is the first work that introduces conditional normalizing flows into the meta-learning framework for few-shot learning.

\section{Methodology} 
\label{sec:method}
In this section, we present our methodology for learning kernels with random Fourier features under the meta-learning framework with limited labels. In Section~\ref{MLK}, we describe the base-learner based on kernel ridge regression. We introduce kernel learning with random features by formulating it as a variational inference problem in Section~\ref{metavrf}. We describe the context inference to leverage the shared knowledge provided by related tasks in Section~\ref{contextinference}. We further enrich  the variational random features by conditional normalizing flows in Section~\ref{MetaVRF-CNF}.

\subsection{Meta-Learning with Kernels}
\label{MLK}
We adopt the episodic training strategy~\cite{ravi2017optimization} commonly used for few-shot meta-learning, which involves \textit{meta-training} and \textit{meta-test} stages. In the \textit{meta-training} stage, a meta-learner is trained to enhance the performance of a base-learner on a \textit{meta-training} set with a batch of few-shot learning tasks, where a task is usually referred to as an episode \cite{ravi2017optimization}. In the \textit{meta-test} stage, the base-learner is evaluated on a \textit{meta-test} set with different classes of data samples from the \textit{meta-training} set.

For the few-shot classification problem, we sample $N$-way $k$-shot classification tasks from the \textit{meta-training} set, where $k$ is the number of labelled examples for each of the $N$ classes. Given the $t$-th task with a support set $\mathcal{S}^{t}=\{(\mathbf{x}_i, \mathbf{y}_i)\}_{i=1}^{N\mathord\times k}$ and query set $\mathcal{Q}^{t}=\{(\tilde{\mathbf{x}}_i, \tilde{\mathbf{y}}_i)\}_{i=1}^m$ ($\mathcal{S}^{t}, \mathcal{Q}^{t} \subseteq \mathcal{X}$), we learn the parameters $\alpha^{t}$ of the predictor $f_{\alpha^{t}}$ using a standard learning algorithm with a kernel trick $\alpha^{t} = \Lambda(\Phi(X), Y)$, where $\mathcal{S}^{t} = \{X, Y\}$.\ Here, $\Lambda$ is the base-learner and $\Phi: \mathcal{X} \rightarrow \mathbb{R}^\mathcal{X}$ is a mapping function from $\mathcal{X}$ to a dot product space $\mathcal{H}$. The similarity measure $\mathtt{k}(\mathbf{x}, \mathbf{x}')=\langle\Phi(\mathbf{x}),\Phi(\mathbf{x}')\rangle$ is called a
kernel~\cite{hofmann2008kernel}.

In traditional supervised learning, the base-learner for the $t$-th single task usually relies on a universal kernel to map the input into a dot product space for efficient learning. Once the base-learner is trained on the support set, its performance is evaluated on the query set using the following loss function:
\begin{equation}
\sum_{(\tilde{\mathbf{x}}, \tilde{\mathbf{y}}) \in \mathcal{Q}^{t}} 
L \left(f_{\alpha^t} \big(\Phi(\tilde{\mathbf{x}} )\big), \tilde{\mathbf{y}}\right),
\end{equation}
where $L(\cdot)$ can be any differentiable function, \eg,~cross-entropy loss. In the meta-learning setting for few-shot learning, we usually consider a batch of tasks.\ Thus, the meta-learner is trained by optimizing the following objective function \textsl{w.r.t.} the empirical loss on $T$ tasks:
\begin{equation}
 \begin{aligned}
\vspace{-3mm}
\sum^T_{t} \sum_{(\tilde{\mathbf{x}}, \tilde{\mathbf{y}} ) \in \mathcal{Q}^{t}} L\left(f_{\alpha^{t}}\big(\Phi^{t}(\tilde{\mathbf{x}})\big), \tilde{\mathbf{y}}\right), \text{s.t.} \,\ \alpha^{t} = \Lambda\left(\Phi^{t}(X), Y\right),
\label{obj}
\vspace{-2mm}
\end{aligned}   
\end{equation}
where $\Phi^t$ is the feature mapping function which can be obtained by learning a task-specific kernel $\mathtt{k}^t$ for each task $t$ with data-driven random Fourier features.

In this work, we employ kernel ridge regression, which has an efficient closed-form solution, as the base-learner $\Lambda$ for few-shot learning.\ The kernel value in the Gram matrix $K \in \mathbb{R}^{Ck\times Ck}$ is computed as $\mathtt{k}(\mathbf{x}, \mathbf{x}') = \Phi(\mathbf{x}) \Phi(\mathbf{x}')^{\top}$, where ``${\top}$'' is the transpose operation. The base-learner $\Lambda$ for a single task is obtained by solving the following objective \textsl{w.r.t.} the support set of this task,
\begin{equation}
\Lambda = \argmin_{\alpha} \Tr[(Y-\alpha K) (Y-\alpha K)^{\top}] + \lambda \Tr[\alpha K \alpha^{\top}],
\label{krg}
\end{equation}
which admits a closed-form solution 
\begin{equation}
\alpha = Y(\lambda \mathrm{I} + K)^{-1}.
\label{closed}
\end{equation}
The learned predictor is then applied to samples in the query set $\tilde{X}$:
\begin{equation}
\hat{Y}=f_{\alpha}(\tilde{X})=\alpha \tilde{K}, 
\end{equation} 
Here, $\tilde{K} = \Phi(X)\Phi(\tilde{X})^\top\in \mathbb{R}^{Ck\times m}$, with each element as $\mathtt{k}(\mathbf{x}, \tilde{\mathbf{x}})$ between the samples from the support and query sets. Note that we also treat $\lambda$ in (\ref{krg}) as a trainable parameter by leveraging the meta-learning setting, and all these parameters are learned by the meta-learner.  

In order to obtain task-specific kernels,  we consider learning adaptive kernels with random Fourier features in a data-driven way. This also enables shared knowledge of different tasks to be captured by exploring their dependencies in the meta-learning framework.

\subsection{Variational Random Features} 
\label{metavrf}

From a probabilistic perspective, under the meta-learning setting for few-shot learning, the random feature basis is obtained by maximizing the conditional predictive log-likelihood of samples from the query set $\mathcal{Q}$: 
\begin{align}
&\max_{p} \sum_{(\mathbf{x},\mathbf{y})\in \mathcal{Q}} \log  p(\mathbf{y} | \mathbf{x}, \mathcal{S}) \\ &= \max_{p} \sum_{(\mathbf{x},\mathbf{y})\in \mathcal{Q}} \log  \int p(\mathbf{y} |\mathbf{x},  \mathcal{S}, \bm{\omega})  p(\bm{\omega} | \mathbf{x},  \mathcal{S}) d\bm{\omega}.
\label{likeli}
\end{align}
We adopt a conditional prior distribution $p(\bm{\omega} | \mathbf{x},  \mathcal{S})$ over the base $\bm{\omega}$, as in the conditional variational autoencoder~\cite{sohn2015learning}, rather than an uninformative prior \cite{kingma2013auto,rezende2014stochastic}. By depending on the input $\mathbf{x}$, we infer the bases that can specifically represent the data, while leveraging the context of the current task by conditioning on the support set $\mathcal{S}$.

In order to infer the posterior $p(\bm{\omega} | \mathbf{y},\mathbf{x}, \mathcal{S})$ over $\bm{\omega}$, which is generally intractable, we use a variational distribution $q_{\phi}(\bm{\omega}| \mathcal{S})$ to approximate it, where the base is conditioned on the support set $\mathcal{S}$ by leveraging meta-learning. We obtain the variational distribution by minimizing the Kullback-Leibler (KL) divergence:

\begin{equation}
\KL[q_{\phi}(\bm{\omega}| \mathcal{S}) || p(\bm{\omega} | \mathbf{y}, \mathbf{x},  \mathcal{S})].
\label{kl}
\end{equation}
By applying  Bayes' rule to the posterior $p(\bm{\omega}|\mathbf{y},\mathbf{x},  \mathcal{S})$, we derive the evidence lower bound (ELBO) as
\begin{align}
\log  p(\mathbf{y} | \mathbf{x},  \mathcal{S}) \geq \,\,\,   &\mathbb{E}_{q_{\phi}(\bm{\omega}| \mathcal{S})} \log \, p(\mathbf{y} | \mathbf{x},  \mathcal{S}, \bm{\omega} ) \nonumber\\ &- \KL[q_{\phi}(\bm{\omega}|\mathcal{S}) || p(\bm{\omega} | \mathbf{x},  \mathcal{S})].
\label{eq:elbo}
\end{align}
The first term of the ELBO is the predictive log-likelihood conditioned on the observation $\mathbf{x}$, $ \mathcal{S}$ and the inferred RFF bases $\bm{\omega}$. Maximizing it enables us to make an accurate prediction for the query set by utilizing the inferred bases from the support set. The second term in the ELBO minimizes the discrepancy between the meta variational distribution $q_{\phi}(\bm{\omega}|\mathcal{S})$ and the meta prior $p(\bm{\omega} | \mathbf{x}, \mathcal{S})$, which encourages samples from the support and query sets to share the same random Fourier bases. The full derivation of the ELBO is provided in the supplementary material.

We now obtain the objective by maximizing the ELBO with respect to a batch of $T$ tasks:
\begin{align}
\vspace{-4mm}
\mathcal{L} = &\frac{1}{T} \sum_{t=1}^{T} \Big[ \sum_{(\mathbf{x},\mathbf{y})\in \mathcal{Q}^{t}} \!\!\!\! \mathbb{E}_{q_{\phi}(\bm{\omega}^t| \mathcal{S}^t)} \log \, p(\mathbf{y} | \mathbf{x},\mathcal{S}^t, \bm{\omega}^t ) \nonumber\\ &- \KL[q_{\phi}(\bm{\omega}^t|\mathcal{S}^t) || p(\bm{\omega}^t | \mathbf{x}, \mathcal{S}^t)]  \Big],
\label{vi-obj-base} 
\end{align}
where $\mathcal{S}^t$ is the support set of the $t$-th task associated with its specific bases $\{\bm{\omega}^t_{d}\}_{d=1}^{D}$ and $(\mathbf{x}, \mathbf{y}) \in \mathcal{Q}^t$ is the sample from the query set of the $t$-th task.


\subsection{Task Context Inference}
\label{contextinference}
\begin{figure}[t]
	\centering
	\includegraphics[width=\linewidth]{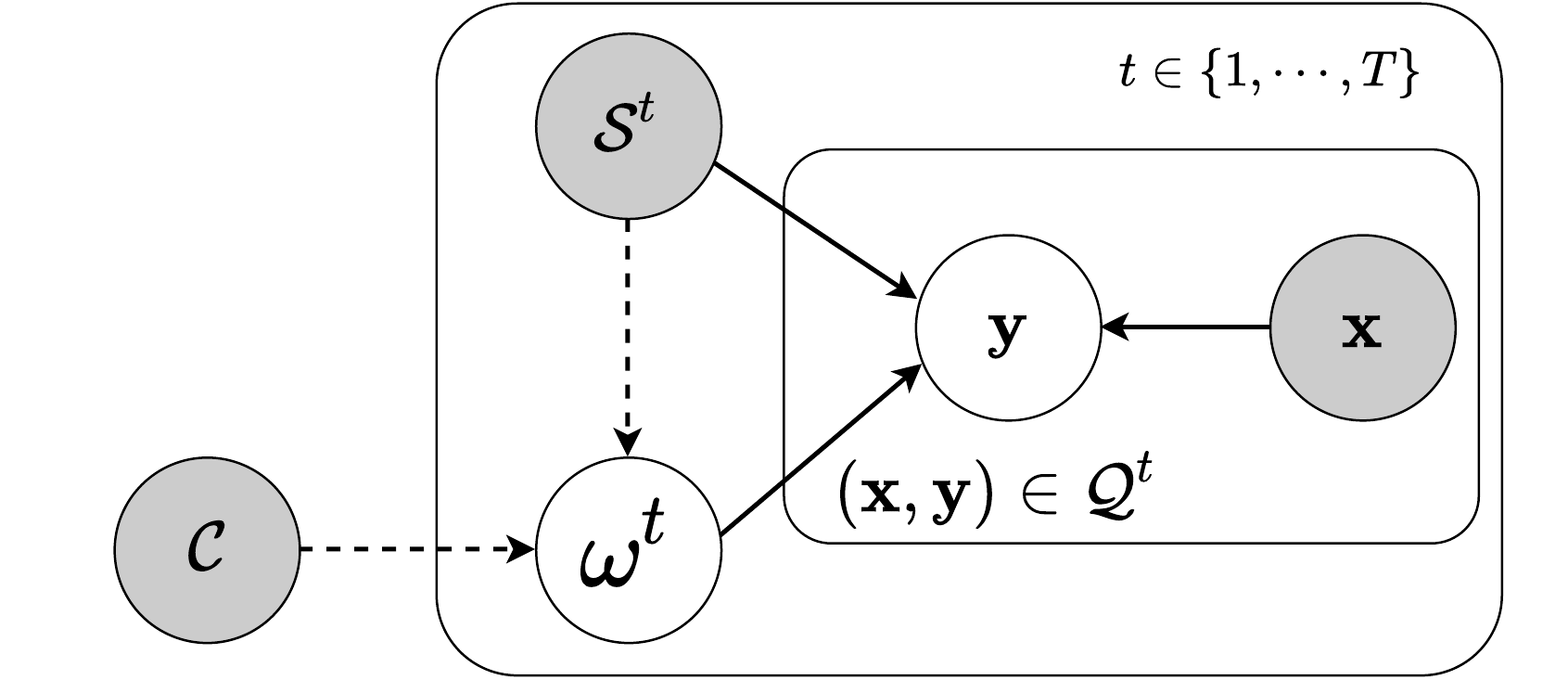}
	\caption{Graphical illustration of variational inference of the random Fourier basis under the meta-learning framework for few-shot learning, where $(\mathbf{x}, \mathbf{y})$ is a sample in the query set $\mathcal{Q}^t$. The base $\bm{\omega}^t$ of the $t$-th task is dependent on the support set $\mathcal{S}^t$ of the current task and the context $\mathcal{C}$ of related tasks. The dashed lines indicate variational inference.}
	\label{graph}
\end{figure}

We propose a context inference which puts the inference of random feature bases for the current task in the context of related tasks. We replace the variational distribution in (\ref{kl}) with a conditional distribution $q_{\phi}(\bm{\omega}^t| \mathcal{S}^t,\mathcal{C})$, where we use $\mathcal{C}$ to contain the shared knoweledge provided by related tasks. This makes the bases $\{\bm{\omega}^t_{d}\}_{d=1}^{D}$ of the current $t$-th task conditioned also on the context $\mathcal{C}$ of related tasks, which gives rise to a new ELBO, as follows:
\begin{equation}
\begin{aligned}
\log  p(\mathbf{y} | \mathbf{x},  \mathcal{S}^t) &\geq \,\,\,   \mathbb{E}_{q_{\phi}(\bm{\omega}| \mathcal{S}^t,\mathcal{C})} \log \, p(\mathbf{y} | \mathbf{x},  \mathcal{S}^t, \bm{\omega} ) \\ &- \KL[q_{\phi}(\bm{\omega}|\mathcal{S}^t,\mathcal{C}) || p(\bm{\omega} | \mathbf{x},  \mathcal{S}^t)].
\label{metaelbo} 
\end{aligned}
\end{equation}
This can be represented in a directed graphical model, as shown in Figure~\ref{graph}. In a practical sense, the KL term in (\ref{metaelbo}) encourages the model to extract useful information from previous tasks for inferring the spectral distribution associated with each individual sample $\mathbf{x}$ of the query set in the current task.

The context inference integrates the knowledge shared across tasks with the task-specific knowledge to build up adaptive kernels for individual tasks. The inferred random features are highly informative due to the  information absorbed from experienced tasks. The base-learner built on the inferred kernel with the informative random features effectively solves the current task.

However, since there is usually a large number of related tasks, it is non-trivial to model them all simultaneously. We consider using recurrent neural networks to gradually accumulate information episodically along with the learning process by organizing tasks in a sequence. We propose an LSTM-based inference network, leveraging its innate capability of remembering long-term information~\cite{gers2000recurrent}. The LSTM offers a well-suited structure to implement the context inference. The cell state $\mathbf{c}$ stores and accrues the meta knowledge shared among related tasks. It can also be updated when experiencing a new task in each episode over the course of learning, where the output $\mathbf{h}$ is used to adapt the model to each specific task. 

To be more specific, we model the variational posterior $q_{\phi}(\bm{\omega}^t| \mathcal{S}^t,\mathcal{C})$ through $q_{\phi}(\bm{\omega}|\mathbf{h}^t)$, which is parameterized as a multi-layer perceptron (MLP) $\phi(\mathbf{h}^t)$. Note that $\mathbf{h}^t$ is the output from an LSTM that takes $\mathcal{S}^t$ and $\mathcal{C}$ as inputs. We implement the inference network with bidirectional LSTMs \cite{schuster1997bidirectional,graves2005framewise}. For the LSTM, we have
\begin{equation}
[\mathbf{h}^t, \mathbf{c}^t] = g_{\mathrm{LSTM}}(\mathcal{\bar{S}}^t,\mathbf{h}^{t-1},\mathbf{c}^{t-1}),
\label{vlstm}
\end{equation}
where $g_{\mathrm{LSTM}}(\cdot)$ is a LSTM network that takes the current support set, the output $\mathbf{h}^{t-1}$ and the cell state $\mathbf{c}^{t-1}$ as input. $\mathcal{\bar{S}}^t$ is the average over the feature representation vectors of samples in the support set~\cite{zaheer2017deep}. The feature representation is obtained by a shared convolutional network $\psi(\cdot)$. To incorporate more context information, we also implement the inference with a bidirectional LSTM. We thus have $\mathbf{h}^t = [\stackrel{\rightarrow}{\mathbf{h}^t}, \stackrel{\leftarrow}{\mathbf{h}^t}]$, 
where $\stackrel{\rightarrow}{\mathbf{h}^t}$ and $\stackrel{\leftarrow}{\mathbf{h}^t}$ are the outputs from the forward and backward LSTMs, respectively, and $[\cdot,\cdot]$ indicates a concatenation operation.

Therefore, the optimization objective with the context inference is:
\begin{equation}
    \begin{aligned}
\mathcal{L} = &\frac{1}{T} \sum_{t=1}^{T} \Big[\sum_{(\mathbf{x},\mathbf{y})\in \mathcal{Q}^{t}} \!\!\!\! \mathbb{E}_{q_{\phi}(\bm{\omega}^t| \mathbf{h}^t)} \log \, p(\mathbf{y} | \mathbf{x},\mathcal{S}^t, \bm{\omega}^t) \\ -& \KL[q_{\phi}(\bm{\omega}^t|\mathbf{h}^t) || p(\bm{\omega}^t | \mathbf{x},\mathcal{S}^t)]  \Big],
\label{vi-obj}
\end{aligned}
\end{equation}
where the variational approximate posterior $q_{\phi}(\bm{\omega}^t| \mathbf{h}^t)$ is taken as a multivariate Gaussian with a diagonal covariance. Given the support set as input, the mean $\bm{\omega}_{\mu}$ and standard deviation $\bm{\omega}_{\sigma}$ are output from the inference network $\phi(\cdot)$. The conditional prior $p(\bm{\omega}^t | \mathbf{x},\mathcal{S}^t)$ is implemented with a prior network which takes an aggregated representation  using the cross attention \cite{kim2019attentive} between $\mathbf{x}$ and $\mathcal{S}^t$. The details of the prior network are provided in the supplementary material. To enable back propagation with the sampling operation during training, we adopt the reparametrization trick \cite{rezende2014stochastic,kingma2013auto} as
$\bm{\omega}= \bm{\omega}_{\mu} + \bm{\omega}_{\sigma} \odot \boldsymbol\epsilon$, where $\bm\epsilon \sim \mathcal{N}(0, \mathrm{I} ).$
%


During the course of learning, the LSTMs accumulate knowledge in the cell state by updating their cells using information extracted from each task. For the current task $t$, the knowledge stored in the cell is combined with the task-specific information from the support set to infer the spectral distribution for this task. To accrue information across all the tasks in the meta-training set, the output and the cell state of the LSTMs are passed down across batches. As a result, the final the cell state contains the distilled prior knowledge from all the tasks experienced in the meta-training set. 


\subsection{Enriching Random Features by Normalizing Flows}
\label{MetaVRF-CNF}
%
%

The posterior distribution $q_{\phi}(\bm{\omega}|\mathbf{h}^t)$ is assumed to be a fully factorized Gaussian, resulting in limited expressive ability to approximate the true posterior over random Fourier bases. Motivated by the empirical success of normalizing flows~\cite{rezende2015variational} and conditional normalizing flows~\cite{winkler2019learning}, we propose the conditional normalizing flows that provide a principled way to learn richer posteriors. 

Normalizing flows map a complex distribution $p_{\mathbf{x}}(\mathrm{X})$ to a simpler distribution $p_{\vect{z}}(\mathrm{Z})$ through a chain of transformations. 
Let $\mathbf{x} \in X$ denote data sampled from an unknown distribution $\mathbf{x} \sim p_{X}(\mathbf{x})$.
The key idea in normalizing flows is to represent $p_{X}(\mathbf{x})$ as a transformation $\mathbf{x}=g(\mathbf{z})$ of a single Gaussian distribution $\mathbf{z} \sim p_{Z} = \mathcal{N}(0, I)$. Moreover, we assume that the mapping is bijective: $\mathbf{x} = g(\mathbf{z}) = f^{-1}(\mathbf{z})$. Therefore, the log-likelihood of the data is given by the change of variable formula:
\begin{equation}
    \begin{aligned}
\label{eq:likelihood}
\log\left(p_X(\mathbf{x})\right) =& \log\left(p_Z\left(f(\mathbf{x})\right)\right)+\log\left( \left|\det\left(\frac{\partial f(\mathbf{x})}{\partial \mathbf{x}^T}\right)\right|\right), 
\end{aligned}
\end{equation}
where $\frac{\partial f(\mathbf{x})}{\partial \mathbf{x}^T}$ is the Jacobian of the map $f(\mathbf{x})$ at $\mathbf{x}$.
The functions $f$ can be learned by maximum likelihood~(\ref{eq:likelihood}), where the bijectivity assumption allows  expressive mappings to be trained by gradient backpropagation. 
\begin{figure}[t]
	\centering
	\includegraphics[width=1\linewidth]{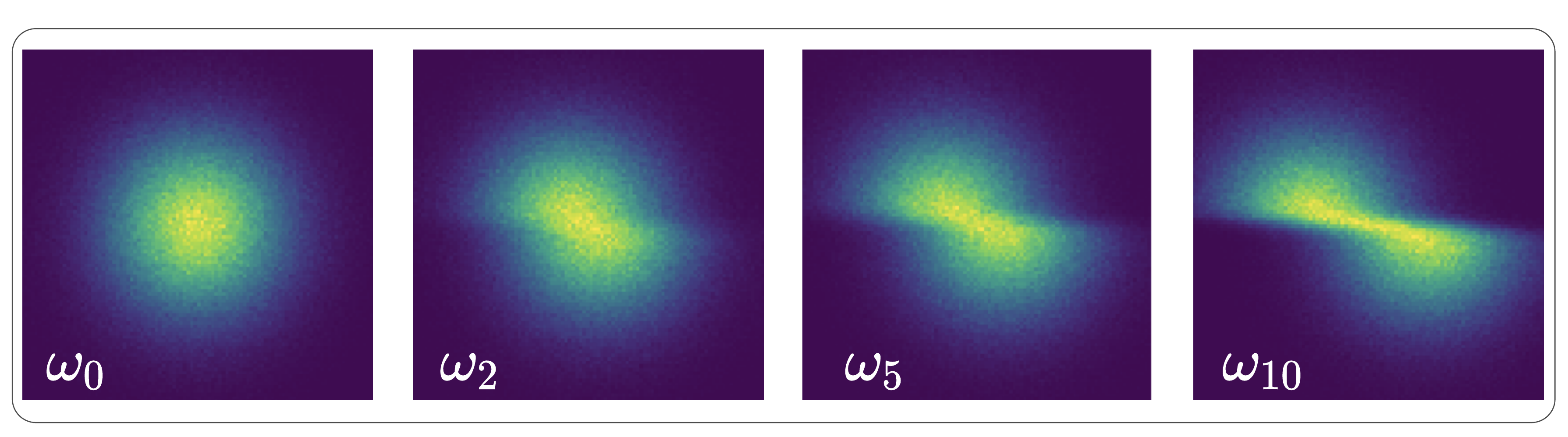}
\label{fig:frame}
    \vspace{-5mm}
	\caption{Effect of conditional normalizing flows on the random bases. They transform the single Gaussian distribution of the random bases into a more complex distribution, which yields more informative random features.}
	\vspace{-3mm}
	\label{fig: nf_distribution}
\end{figure}

To make the Jacobian tractable for the map $f(\mathbf{x})$, NICE~\cite{dinh2014nice} and RealNVP~\cite{dinh2016density} proposed to stack a sequence of simple bijective transformations, such that their Jacobian is a triangular matrix. In this way, the log-determinant depends only on the sum of its diagonal elements. Dinh \etal~\cite{dinh2014nice, dinh2016density} proposed the additive coupling layer for each transformation.  In each affine coupling transformation, the input vector $\mathbf{x}\in \mathbb{R}^d$ is split into upper and lower halves,  $\mathbf{x}_{I_1},\mathbf{x}_{I_2} \in \mathbb{R}^{d/2}$. These are plugged into the following transformation, referred to as a single flow-block  $f_i$:
\begin{equation}
    \begin{aligned}\label{eq:3}
\mathbf{z}_1 = \mathbf{x}_{I_1},~~~ \mathbf{z}_2 = \mathbf{x}_{I_2} \circ \exp(s_i(\mathbf{x}_{I_1})) + t_i(\mathbf{x}_{I_1}), 
\end{aligned}
\end{equation}
where $\circ$ denotes element-wise multiplication.  It is important to note that the mappings $s_i$ and $t_i$ can be arbitrarily complicated functions of $\mathbf{x}_i$ and need not be invertible themselves. In practice, $s_i$ and $t_i$ are achieved via neural networks. 

Given the outputs $\mathbf{z}_1$ and $\mathbf{z}_2$, this affine transformation is  invertible by:
\begin{equation}
    \begin{aligned}\label{eq:4}
\mathbf{x}_{I_1} = \mathbf{z}_1,~~~ \mathbf{x}_{I_2} = (\mathbf{z}_2 - t_i(\mathbf{z}_1)) \circ \exp(-s_i(\mathbf{z}_1)). 
\end{aligned}
\end{equation}

The RealNVP~\cite{dinh2016density} flow comprises $k$ reversible flow-blocks interleaved with switch-permutations,
\begin{equation}
    f_{\textit{RealNVP}} = f_k\cdot r \dots f_2 \cdot r \cdot f_1, 
\end{equation}
where $r$ denotes a switch-permutation, which permutes the order of $\mathbf{x}_1$ and $\mathbf{x}_2$. According to the chain rule, the log-determinant of the Jacobian of the whole transformation $f$ is computed by summing the  log-determinants of the Jacobian of each $f_i$, making the likelihood calculation tractable.

Conditional normalizing flows~\cite{winkler2019learning} learn conditional likelihoods for complicated target distributions in multivariate prediction tasks.
Take an input $\mathbf{x} \in \mathcal{X}$ and a regression target $\mathbf{y} \in \mathcal{Y}$. CNFs learn a complicated distribution $p_{Y|X}(\mathbf{y} | \mathbf{x})$ using a conditional prior $p_{Z|X}(\mathbf{z} | \mathbf{x})$ and a mapping $f_\phi: \gY \times \gX \to \gZ$, which is bijective in $\gY$ and $\gZ$. The log-likelihood of CNFs is:
\begin{equation}
\begin{aligned}
     \log (p_{Y|X}(\mathbf{y} | \mathbf{x})) = & \log(p_{Z|X}(\mathbf{z} | \mathbf{x})) + \log(\left\lvert \frac{\partial \mathbf{z}}{\partial \mathbf{y}} \right\rvert) \\ = & \log(p_{Z|X}(f_{\phi}(\mathbf{y} , \mathbf{x}) | \mathbf{x})) + \log(\left\lvert \frac{\partial f_{\phi}(\mathbf{y} , \mathbf{x})}{\partial \mathbf{y}} \right\rvert). \label{eq:cnf} 
\end{aligned}
\end{equation}
Different from NFs, in the log-likelihood of CNFs, all distributions are conditional and the flow has a conditioning argument for $\mathbf{x}$.


We parameterize the approximate posterior distribution $q_{\phi}(\bm{\omega}|\mathbf{h}^t)$ with a flow of length $K$, $q_{\phi}(\bm{\omega}|\mathbf{h}^t) := q_{K}(\bm{\omega}_K)$.
The ELBO~(\ref{eq:elbo}) is thus written as an expectation over the initial distribution $q_0(\bm{\omega})$:
\begin{equation}
\begin{aligned}
  \log  p(\mathbf{y} | \mathbf{x},  \mathcal{S}) \geq & -\E_{q_{\phi} ( \bm{\omega} |\mathbf{h}^t )} [ \log q_{\phi}(\bm{\omega}|\mathbf{h}^t) + \log  p ( \mathbf{y}, \bm{\omega} | \mathcal{S}, \mathbf{x} ) ] \\
  = &-\E_{q_{0} ( \bm{\omega}_0 )} \left [ \ln  q_{K} ( \bm{\omega}_K ) + \log   p ( \mathbf{y} ,\bm{\omega}_K|\mathcal{S}, \mathbf{x}) \right ] \\
  =  &-\E_{q_{0} (\bm{\omega}_0)} [ \ln  q_{0} ( \bm{\omega}_0 ) - \sum_{k=1}^{K} \ln |\det \frac{\partial f}{\partial \bm{\omega}_k}| ] \\
  &  + \E_{q_{0} ( \bm{\omega}_0 )} [\log  p ( \mathbf{y} ,\bm{\omega}_K|\mathcal{S}, \mathbf{x}) ], 
  \label{eq:elbo-nf}
\end{aligned}
\end{equation}
where $q_0(\bm{\omega}_0)$ is obtained from the approximate posterior distribution $q_{\phi}(\bm{\omega}|\mathbf{h}^t)$ without transformation.

We then obtain the objective by maximizing the log-likelihood $ \log  p(\mathbf{y} | \mathbf{x},  \mathcal{S})$ with respect to a batch of $T$ tasks:
\begin{equation}
    \begin{aligned}
      \mathcal{L} = &\frac{1}{T} \sum_{t=1}^{T} \Big[ \sum_{(\mathbf{x},\mathbf{y})\in \mathcal{Q}^{t}} \!\!\!\! \E_{q_{0} (\bm{\omega}_0^t)} [ - \ln  q_{0} ( \bm{\omega}_0^t ) + \sum_{k=1}^{K} \ln |\det \frac{\partial f}{\partial \bm{\omega}_k^t}| ] \\
   +&  \E_{q_{0} ( \bm{\omega}_0^t )} \left[\log  p ( \mathbf{y} ,\bm{\omega}^t_K|\mathcal{S}^t, \mathbf{x}) \right] \Big], 
     \label{eq:obj-cnf}
    \end{aligned}
\end{equation}
where $\bm{\omega}_k^t$ is the random base after $k$ transformations.

We rely on the conditional coupling layer from~\cite{winkler2019learning} to transform the random base distribution. This layer is an extension of the affine coupling layer from RealNVP~\cite{dinh2016density}  to make the computation of the Jacobian for the map $f(x)$ tractable. The input $\bm{\omega}_{k-1} = [\bm{\omega}_{k-1}^{I_0}, \bm{\omega}_{k-1}^{I_1}]$ of an affine coupling layer is split into two parts, which are transformed individually: 
\begin{equation}
    \begin{aligned}
    	\bm{\omega}_{k}^{I_i} = \bm{\omega}_{k-1}^{I_i} \odot  \exp(s_{i+1}(\bm{\omega}_{k-(1-i)}^{I_{(1-i)}}, \mathbf{h}^t)& \\
    	+ t_{(i+1)}(\bm{\omega}_{k-(1-i)}^{I_{(1-i)}}, \mathbf{h}^t)& 
    \end{aligned}
\end{equation}
where $i \in \{0, 1\}$. Note that the transformations $s_{i+1}, t_{i+1}$ do not need to be invertible and are modelled as convolutional neural networks. The inverse of an affine coupling layer is:
\begin{equation}
    \begin{aligned}
	\bm{\omega}_{k-1}^{I_i} = (\bm{\omega}_{k}^{I_i} - t_{(1+i)}(\bm{\omega}_{k-(1-i)}^{I_1}, \mathbf{h}^t))&\\ \odot \exp(-s_{(1+i)}(\bm{\omega}_{k-i}^{I_i}, \mathbf{h}^t))&.
    \end{aligned}
\end{equation}
The log-determinant of the Jacobian for one affine coupling layer is calculated as the sum over $s_i$, \ie, $\sum_j s_1(\bm{\omega}_{k-1}^{I_1}, \mathbf{h}^t)_j + \sum_j s_2(\bm{\omega}_{k}^{I_0}, \mathbf{h}^t)_j$. 
%
A deep invertible network is built as a sequence of multiple such layers, with a permutation of the dimensions after each layer. 
The conditional input $\mathbf{h}^t$ is added as an extra input to each transformation in the coupling layer. We refer to the kernel constructed based on the random bases by conditional normalizing flows as MetaKernel. 

We visualize the distribution of the random bases produced by the CNFs in Figure~\ref{fig: nf_distribution}. $\bm{\omega}_k$ indicates the distribution of the random bases after $k$ transformations. This visualization shows that we can transform a single Gaussian distribution of random bases into a more complex distribution, which  achieves more informative random features, resulting in improved performance, as we will demonstrate in our experiments. 

\section{Experiments} 
\label{sec:experiments}
In this section, we report our experiments to demonstrate the effectiveness of the proposed MetaKernel for both regression and classification with limited labels. We also provide thorough ablation studies to gain insight into our method by showing the efficacy of each introduced component.

\subsection{Few-Shot Classification}
\begin{figure*}[t]
	\centering
	\includegraphics[width=1.\linewidth]{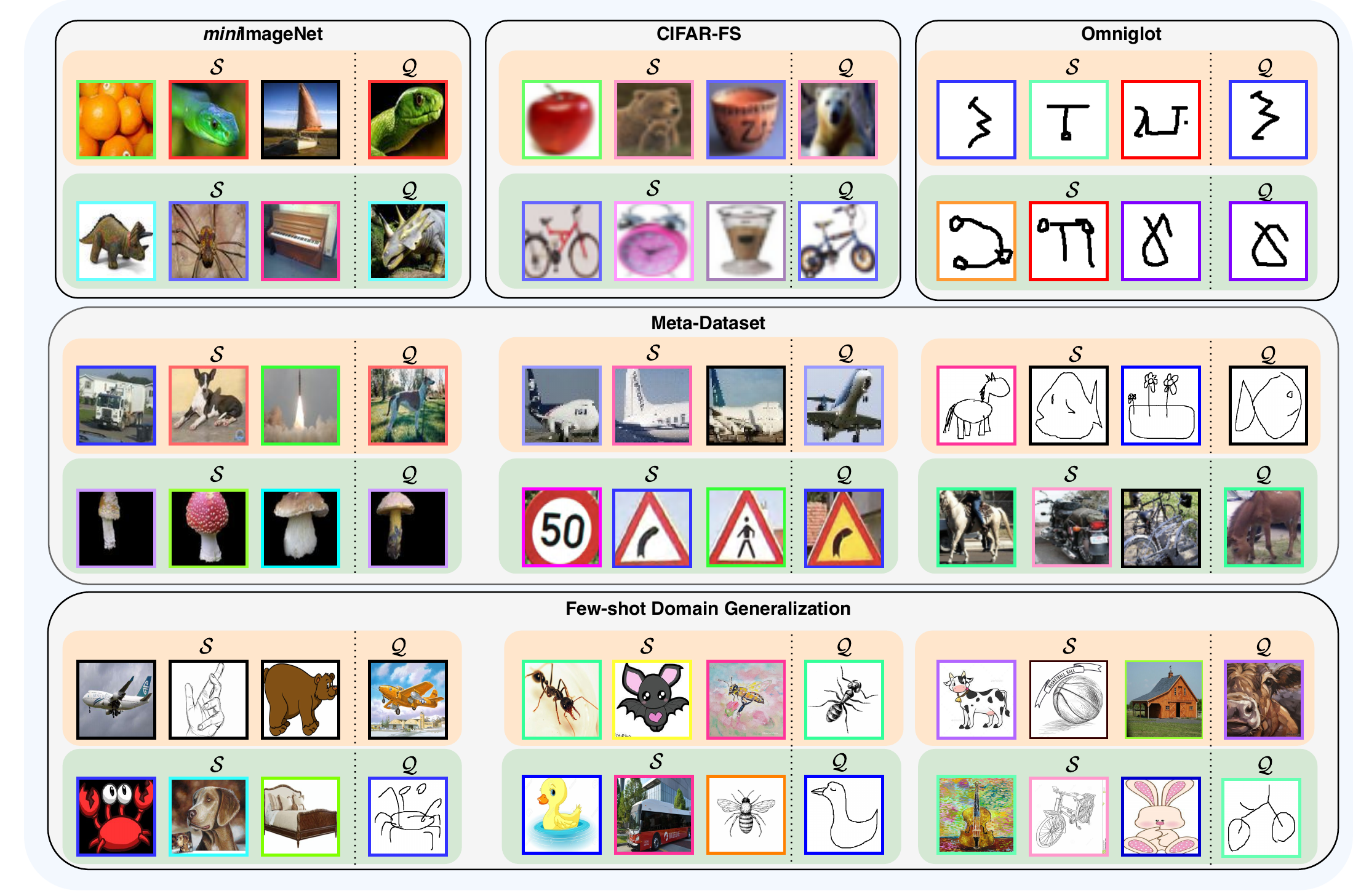}
	\caption{Examples from each dataset. Orange and green boxes indicate the meta-training and meta-test tasks for each dataset. $\mathcal{S}$ and $\mathcal{Q}$ indicate the support and query sets for each task. For Meta-Dataset, we only show examples from \textit{ImageNet}~\cite{russakovsky2015imagenet}, \textit{Aircraft}~\cite{maji13finegrained}, \textit{Quick Draw}~\cite{Quick}, \textit{Fungi}~\cite{Fungi},\textit{Traffic Signs}~\cite{Houben-IJCNN-2013} and
\textit{MS-COCO}~\cite{lin2014microsoft}. For the few-shot domain generalization, we only show the examples from DomainNet using \textit{Quick Draw} as the target domain during the meta-test stage.} 
	\label{fig:Dataset}
\end{figure*}

The few-shot classification experiments are conducted on four commonly used benchmarks, \ie, Omniglot \cite{lake2015human}, \mini{} \cite{vinyals2016matching}, CIFAR-FS \cite{krizhevsky2009learning} and Meta-Dataset~\cite{triantafillou2019meta}.
We also perform experiments on  DomainNet~\cite{peng2019moment} for few-shot domain generalization. Sample images from each dataset are provided in Figure~\ref{fig:Dataset}.

\subsubsection{Datasets}
\textbf{Omniglot}~\cite{lake2015human} is a few-shot classification benchmark that contains $1623$ handwritten characters (each with $20$ examples). All characters are grouped into one of $50$ alphabets. For fair comparison against the state of the art, we follow the same data split and pre-processing used by Vinyals \etal~\cite{vinyals2016matching}. Specifically, the training, validation, and test sets are composed of a random split of $[1100, 200, 423]$. The dataset is augmented with rotations of $90$ degrees, which results in $4000$ classes for training, $400$ for validation, and $1292$ for testing. The number of examples is fixed to $20$. All images are resized to $28\mathord\times 28$. For a $N$-way, $k$-shot task at training time, we randomly sample $N$ classes from the $4000$ classes, each with $(k+15)$ examples. Thus, there are $C\mathord\times k$ examples in the support set and $C \mathord\times 15$ examples in the query set. The same sampling strategy is followed for validation and testing.

\textbf{\mini{}}~\cite{vinyals2016matching} is a challenging dataset constructed from ImageNet~\cite{russakovsky2015imagenet}, which comprises a total of  $100$ different classes (each with $600$ instances). All  images are downsampled to $84\mathord\times 84$.
We use the same splits as Ravi and Larochelle~\cite{ravi2017optimization}, with $[64, 16, 20]$ classes for training, validation and testing.  We use the same episodic sampling strategy as for Omniglot.

\textbf{\cifarfs{}}~\cite{bertinetto2018meta} is adapted from CIFAR-100~\cite{krizhevsky2009learning} for few-shot learning.  In the many-shot image classification benchmark CIFAR-100, there are $100$ classes grouped into $20$ superclasses (each with $600$ instances). \cifarfs{} uses the same split criteria ($64, 16, 20$) with which \mini{} has been generated. The resolution of all images is $32\mathord\times 32$.

\textbf{Meta-Dataset} \cite{triantafillou2019meta} is composed of ten  existing image classification datasets (eight for training, two for testing). These are: \textit{ILSVRC-2012} (ImageNet, \cite{russakovsky2015imagenet}), \textit{Omniglot}~\cite{lake2015human}, \textit{Aircraft}~\cite{maji13finegrained}, \textit{CUB-200-2011}
(Birds, \cite{WahCUB_200_2011}), \textit{Describable Textures}~\cite{cimpoi14describing},
\textit{Quick Draw}~\cite{Quick}, \textit{Fungi}~\cite{Fungi}, \textit{VGG Flowr}~\cite{Nilsback08}, \textit{Traffic Signs}~\cite{Houben-IJCNN-2013} and
\textit{MS-COCO}~\cite{lin2014microsoft}.  Each episode generated in Meta-Dataset uses classes from a single dataset. 
Two of these datasets, \textit{Traffic Signs} and \textit{MSCOCO}, are fully reserved for evaluation, which means that no classes from these sets are participated in the training set. 
Apart from for \textit{Traffic Signs} and \textit{MS-COCO}, 
the remaining datasets contribute some classes to the training, validation and test splits. There are about 14 million images in total in Meta-Dataset. 

\textbf{DomainNet.}~\cite{peng2019moment}. Du \etal~\cite{du2020metanorm} introduced the setting of few-shot domain generalization, which combines the challenges of both few-shot classification and domain generalization. It is based on the DomainNet dataset by Peng \etal~\cite{peng2019moment}, which contains six distinct domains, \ie, \textit{clipart}, \textit{infograph}, \textit{painting}, \textit{quickdraw}, \textit{real}, and \textit{sketch}, for 345 categories. The categories are from 24 divisions.

\renewcommand{\arraystretch}{1.3}
\begin{table*}[t]
\caption{Few-shot image classification performance ($\%$) on \mini{}, \cifarfs{}, and Omniglot. Best performing methods and any other runs within the $96\%$ confidence margin are obtained in bold. MetaKernel consistently achieves the top performance.} 
\centering
\begin{threeparttable}
\begin{tabular}{lllllllll}
\toprule
& \multicolumn{2}{c}{\textbf{\mini{}, 5-way}} & \multicolumn{2}{c}{\textbf{\cifarfs{}, 5-way}} & \multicolumn{2}{c}{\textbf{Omniglot, 5-way}} & \multicolumn{2}{c}{\textbf{Omniglot, 20-way}} \\
\cmidrule(lr){2-3} \cmidrule(lr){4-5} \cmidrule(lr){6-7} \cmidrule(lr){8-9}
\textbf{Method} & 1-shot & 5-shot & 1-shot & 5-shot & 1-shot & 5-shot & 1-shot & 5-shot\\
\midrule
\textbf{\textsc{Meta-LSTM}}~\cite{ravi2017optimization} & 43.4$\pm$0.8  & 60.6$\pm$0.7  & --- & --- & --- & --- & --- & ---  \\
\textbf{\textsc{Matching net}}~\cite{vinyals2016matching} & 44.2  & 57  & --- & ---  & 98.1  & 98.9  & 93.8  & 98.5 \\
\textbf{\textsc{SNAIL}} (32C)~by \cite{bertinetto2018meta} & 45.1  & 55.2  & --- & --- & 99.1$\pm$0.2  & 99.8 $\pm$0.1  & 97.6 $\pm$0.3  & \textbf{99.4} $\pm$0.2\\
\textbf{\textsc{MAML}} ($64$C) & 46.7$\pm$1.7  & 61.1$\pm$0.1  & 58.9$\pm$1.8  & 71.5$\pm$1.1  & ---  & --- & --- & ---\\
\textbf{\textsc{MAML}}~\cite{finn2017model} & 48.7$\pm$1.8  & 63.1$\pm$0.9  & 58.9$\pm$1.9  & 71.5$\pm$1.0   & 98.7$\pm0.4$  & \textbf{99.9}$\pm$0.1  & 95.8$\pm$0.3  & 98.9$\pm$0.2 \\
\textbf{\textsc{Protonet}}~\cite{snell2017prototypical} & 47.4$\pm$0.6  & 65.4$\pm$0.5  & 55.5$\pm$0.7  & 72.0$\pm$0.6 & 98.5$\pm$0.2  & 99.5$\pm$0.1  & 95.3$\pm$0.2  & 98.7$\pm$0.1 \\
\textbf{\textsc{iMAML}}~\cite{rajeswaran2019meta} & 49.3$\pm$1.9  &--- & --- & ---  & --- & --- & --- & ---\\
\textbf{R2-D2} ($64$C)~\cite{bertinetto2018meta} & 49.5$\pm$0.2  & 65.4$\pm$0.2  & 62.3$\pm$0.2  & \textbf{77.4}$\pm$0.2  & ---  & --- & --- & ---\\
\textbf{\textsc{OVEPGGP~\cite{snell2020bayesian}}} & 50.0$\pm$0.4 & 67.1$\pm$0.2 & --- & --- & --- & --- & --- & --- \\
\textbf{\textsc{PLATIPUS}}~\cite{finn2018probabilistic} & 50.1$\pm$1.9  & --- & --- & ---  & --- & --- & --- & ---\\
\textbf{\textsc{GNN}}~\cite{garcia2018few} & 50.3  & 66.4  & 61.9  & 75.3  & 99.2  & 99.7  & 97.4  & 99.0\\
\textbf{\textsc{Relation net}}~\cite{sung2018learning} & 50.4$\pm$0.8  & 65.3$\pm$0.7  & 55.0$\pm$1.0  & 69.3$\pm$0.8 & 90.6 $\pm$0.2  & \textbf{99.8}$\pm$0.1  & 97.6$\pm$0.2  & 99.1$\pm$0.1 \\
\textbf{\textsc{R2-D2}}~\cite{devosreproducing} & 51.7$\pm$1.8  & 63.3$\pm$0.9  & 60.2$\pm$1.8  & 70.9$\pm$0.9 & 98.6  & 99.7 & 94.7  & 98.9  \\
\textbf{\textsc{CAVIA}}~\cite{zintgraf2019fast} & 51.8$\pm$0.7  & 65.6$\pm$0.6  & --- & --- & --- & --- & --- & --- \\
\textbf{\textsc{VERSA}}~\cite{gordon2018meta} & \textbf{53.3}$\pm$1.8  & \textbf{67.3}$\pm$0.9  & \textbf{62.5}$\pm$1.7  & 75.1$\pm$0.9  & \textbf{99.7}$\pm$0.2  & \textbf{99.8}$\pm$0.1  & 97.7$\pm$0.3  & 98.8$\pm$0.2  \\
\textbf{\textsc{MetaVRF}}~\cite{zhen2020learning} & \textbf{54.2}$\pm$0.8  & \textbf{67.8}$\pm$0.7  & \textbf{63.1}$\pm$0.7  & \textbf{76.5}$\pm$0.9  &  \textbf{99.8}$\pm$0.1  & \textbf{99.9}$\pm$0.1 &  97.8$\pm$0.3   & \textbf{99.2}$\pm$0.2 \\
\rowcolor{Gray}
\textbf{\textsc{MetaKernel}}  & \textbf{55.5}$\pm$0.9  & \textbf{68.5}$\pm$0.8  & \textbf{64.3}$\pm$0.8  & \textbf{77.5}$\pm$0.9  & \textbf{99.9}$\pm$0.1  & \textbf{99.9}$\pm$0.1  & \textbf{98.7}$\pm$0.3   & \textbf{99.6}$\pm$0.2  \\
\bottomrule
\end{tabular}
\end{threeparttable}
\label{tab:miniandcifar}
\end{table*}

\subsubsection{Implementation Details}
We extract image features using a shallow convolutional neural network with the same architecture as~\cite{gordon2018meta} for \mini{}, and \cifarfs{}. We do not use any fully connected layers in this CNNs. 
For the Meta-Dataset experiments, we use a ResNet-18~\cite{resnet} as our base learner to be consistent  with~\cite{triantafillou2019meta}. The dimension of all feature vectors is $256$. We also evaluate the random Fourier features (RFFs) and the radial basis function (RBF) kernel, where we take the bandwidth $\sigma$ as the mean of the pair-wise distances between samples in the support set of each task. The inference network $\phi(\cdot)$ is a three-layer MLP with $256$ units in the hidden layers and rectifier non-linearity, 
where the input sizes is $512$ for the bidirectional LSTMs.
We use an SGD optimizer with a momentum of $0.9$ in all experiments.


\renewcommand{\arraystretch}{1.3}
\begin{table*}
\caption{Few-shot Meta-Dataset classification accuracy (\%) with variable number of ways and shots, following the setup in \cite{triantafillou2019meta}. 1000 tasks are sampled for evaluation. MetaKernel is a consistent top-performer.}
\centering
{%
\resizebox{1\linewidth}{!}{
\begin{tabular}{lcccccccc}
\toprule
 \multicolumn{1}{l}{\textbf{Dataset}} &
  \multicolumn{1}{c}{\textbf{Matching Net}~\cite{vinyals2016matching}} &
   \multicolumn{1}{c}{\textbf{ProtoNet}~\cite{snell2017prototypical}} &
    \multicolumn{1}{c}{\textbf{fo-MAML}~\cite{finn2017model}} &
     \multicolumn{1}{c}{\textbf{Relation Net}~\cite{sung2018learning}} &
 \multicolumn{1}{c}{\textbf{fo-Proto-MAML}~\cite{triantafillou2019meta}} &
 \multicolumn{2}{c}{\textbf{RFS}~\cite{tian2020rethinking}} &
 \multicolumn{1}{c}{\textbf{MetaKernel}}
 \\
 \cmidrule(lr){7-8}
 & & & & & & LR-Simple  & LR-Distill & \\
\midrule
ILSVRC &  $45.00$ & $50.50$ & $45.51$ &$34.69$ & $49.53$ & $60.14$ &{$61.48$} &{$\textbf{61.71}$} \\
Omniglot & $52.27$ & $59.98$ & $55.55$ & $45.35$ & $63.37$ & $64.92$ & $64.31$ &$\textbf{65.43}$ \\
Aircraft & $48.97$ & $53.10$ & $56.24$ & $40.73$ & $55.95$ &{$63.12$} & $62.32$ &{$\textbf{65.37}$} \\
Birds & $62.21$ & $68.79$ & $63.61$ & $49.51$ & $68.66$ & $77.69$ &{$\textbf{79.47}$} &{$77.13$} \\ 
Textures & $64.15$ & $66.56$ & $68.04$ & $52.97$ & $66.49$ & $78.59$ &{$79.28$} & $\textbf{82.01}$ \\ 
Quick Draw & $42.87$ & $48.96$ & $43.96$ & $43.30$ & $51.52$ &{$\textbf{62.48}$} &{$60.83$}  & $58.46$\\ 
Fungi & $33.97$ & $39.71$ & $32.10$ & $30.55$ & $39.96$ & $47.12$ &{$48.53$} &{$\textbf{49.73}$}\\ 
VGG Flower & $80.13$ & $85.27$ & $81.74$ & $68.76$ & $87.15$ &{$91.60$} & $91.00$  &{$\textbf{93.16}$}\\ 
Traffic Signs & $47.80$ & $47.12$ & $50.93$ & $33.67$ & $48.83$ &{$77.51$} & $76.33$   &{$\textbf{77.91}$} \\ 
MSCOCO & $34.99$ & $41.00$ & $35.30$ & $29.15$ & $43.74$ &{$57.00$} &{$\textbf{59.28}$}  & $56.97$\\\hline
\textit{Mean Accuracy} & $51.24$ & $56.10$ & $53.30$ & $42.87$ & $57.52$ & $68.02$ &{$68.28$}  & {$\textbf{68.79}$}\\
\bottomrule
\end{tabular}}}
\label{tab:meta_dataset}
\end{table*}

The key hyperparameter for the number of bases $D$ in (\ref{rfs}) is set to $D{=}780$ for MetaKernel in all experiments, while we use RFFs with $D{=}2048$ as this produces the best performance. The sampling rate in MetaKernel is much lower than in previous works using RFFs, in which $D$ is usually set to be $5$ to $10$ times the dimension of the input features~\cite{yu2016orthogonal, rahimi2008random}. We adopt a similar meta-testing protocol as~\cite{gordon2018meta, finn2017model}, but we test on $3000$ episodes rather than $600$ and present the results with $95\%$ confidence intervals. All reported results are produced by models trained from scratch. We compare with previous methods that use the same training procedures and similar shallow conventional CNN architectures as ours. Our code will be publicly released.

\renewcommand{\arraystretch}{1.3}\begin{table}[t]
\caption{Few-shot domain generalization performance (5-way $\%$). The best performing methods and any other runs within $95\%$ confidence margin are denoted in bold. MetaKernel is on par with MetaNorm for 1-shot and outperforms all previous methods for the 5-shot setting.}\label{tab:fewdg}
\centering
\begin{tabular}{lcc}
\toprule
\textbf{Method} & 1-shot & 5-shot \\
\midrule
ProtoNets~\cite{snell2017prototypical}& 28.4$\pm$1.8& 47.9$\pm$0.8  \\
MAML~\cite{finn2017model} & 28.7$\pm$1.8& 49.3$\pm$0.8  \\
VERSA~\cite{gordon2018meta}& 30.9$\pm$1.7& 51.7$\pm$0.8  \\
MetaNorm~\cite{du2020metanorm}& \textbf{32.7}$\pm$1.7& 51.9$\pm$0.8  \\

\rowcolor{Gray}
\textbf{\textsc{MetaKernel}} & \textbf{34.7}$\pm$1.7  & \textbf{53.7}$\pm$0.8 \\ 
\bottomrule
\end{tabular}
\end{table}

\begin{figure*}[t]
	\centerline{\includegraphics[width=1\linewidth]{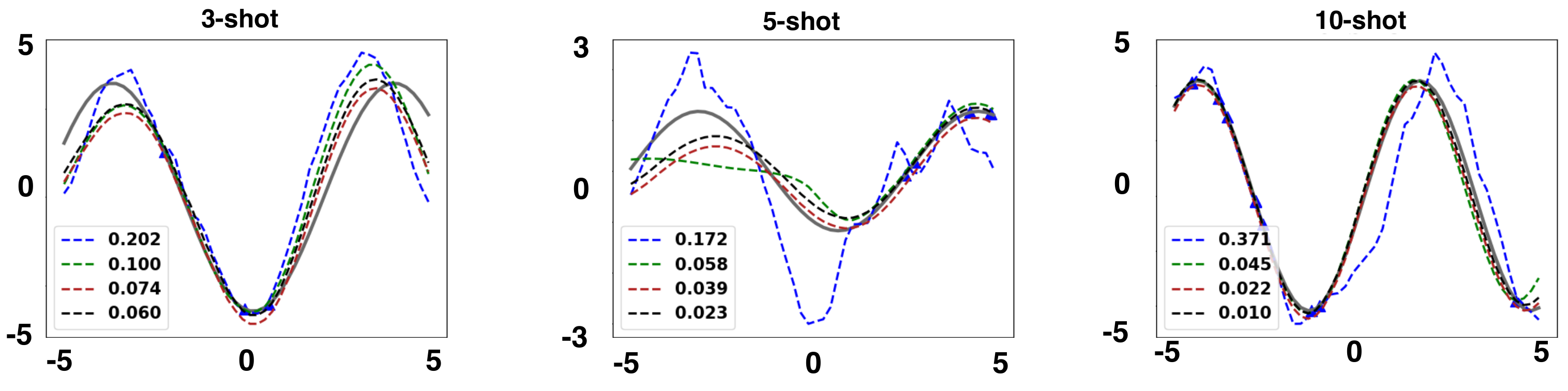}}
	\sbox1{\blueline} \sbox2{\greenline} \sbox3{\redline} \sbox4{\blackline} \sbox5{\grayline} \sbox6{\purplerectangle}
	\caption{
	Few-shot regression
	performance comparison (MSE). MetaKernel fits the target function well, even with variational random features only using three shots, and consistently outperforms MAML for all settings.
	Legend: \usebox1 MAML; \usebox2 MetaKernel (variational RFFs only); \usebox3 MetaKernel~(variational RFFs \& task context); \usebox4 MetaKernel (full model);	\usebox5 Ground Truth; 
	\usebox6 Support Samples.
	}
\label{fig:reg} 
\end{figure*}

\renewcommand{\arraystretch}{1.3}
\begin{table*}[t]
\caption{Ablation studies on \mini{} and \cifarfs{} demonstrating benefit of random Fourier features, task context inference and feature enriching by normalizing flows. Best settings within the 95\% confidece margin are denoted in bold.
}\label{tab:mini_kernel}
\centering
\begin{tabular}{lllll}
\toprule
& \multicolumn{2}{c}{\textbf{\mini{}, 5-way}} & \multicolumn{2}{c}{\textbf{\cifarfs{}, 5-way}} \\
\cmidrule(lr){2-3} \cmidrule(lr){4-5}
\textbf{Method} & 1-shot & 5-shot & 1-shot & 5-shot \\
\midrule
RBF Kernel & 42.1$\pm$1.2  & 54.9$\pm$1.1  &46.0$\pm$1.2  & 59.8$\pm$1.0  \\
RFFs   & 52.8$\pm$0.9  & 65.4$\pm$0.9  & 61.1$\pm$0.8  & 74.7$\pm$0.9  \\
Variational RFFs & 51.3$\pm$0.8  & 66.1$\pm$0.7  & 61.1$\pm$0.7  & 74.3$\pm$0.9  \\
Variational RFFs \& task context inference & \textbf{54.2}$\pm$0.8  & \textbf{67.8}$\pm$0.7  & \textbf{63.1}$\pm$0.7  & \textbf{76.5}$\pm$0.9  \\
\rowcolor{Gray}
\textbf{\textsc{MetaKernel}}  & \textbf{55.5}$\pm$0.9  & \textbf{68.5}$\pm$0.8  & \textbf{64.3}$\pm$0.8  & \textbf{77.5}$\pm$0.9  \\
\bottomrule
\end{tabular}
\end{table*}

\subsubsection{Comparison to the State of the art}
\textbf{Few-shot image classification.}
We first evaluate MetaKernel on the \mini{}, \cifarfs{} and  Omniglot datasets under various way (the number of classes used in each task) and shot (the number of support set examples used per class) configurations. 
The results are reported in Table~\ref{tab:miniandcifar}. 
We report the results of two experiments using MAML~\cite{finn2017model}.  
To keep  MAML~\cite{finn2017model} consistent with our backbone for  \mini{} and \cifarfs{}, in addition to its original results, we also implement  MAML ($64C$) with $64$ channels in each convolutional layer for fair comparison. 
%
While it obtains modest performance, we believe the increased model size leads to overfitting. 
As the original SNAIL uses a very deep ResNet-12 network for embedding, we cite the results of SNAIL reported in \cite{bertinetto2018meta} using a similar shallow network as ours. For fair comparison, we also cite the original results of R2-D2~\cite{bertinetto2018meta} using $64$ channels. 
On all benchmark datasets, MetaKernel delivers the best performance. 
It is worth noting that MetaKernel achieves an accuracy of $55.5\%$ under the $5$-way $1$-shot setting on the \mini~dataset, 
%
surpassing the second-best model by $1.3\%$. This is a good improvement considering the challenge of this setting.
On \cifarfs{}, our model surpasses the second-best method, \ie, VERSA~\cite{gordon2018meta} and has a smaller margin of error bar under the  $5$-way $1$-shot setting using the same backbone. 
On Omniglot, performance of all methods saturates. Nonetheless, MetaKernel achieves the best performance under most settings, including $5$-way $1$-shot, $5$-way $5$-shot, and $20$-way $1$-shot. It is also competitive under the $20$-way $5$-shot setting, falling within the error bars of the state of the art. 

\textbf{Few-shot meta-dataset classification.}
Next, we evaluate MetaKernel on the most challenging few-shot classification benchmark \ie,~Meta-Dataset~\cite{triantafillou2019meta}, which is composed of 10 image classification datasets. 
For Meta-Dataset, we train our model on the ILSVRC~\cite{russakovsky2015imagenet} training split and test on the 10 diverse datasets. As shown in Table~\ref{tab:meta_dataset}, MetaKernel outperforms fo-Proto-MAML~\cite{triantafillou2019meta} across all 10 datasets. MetaKernel also surpasses the second-best method, RFS~\cite{tian2020rethinking}, on 7 out of 10 datasets. Overall, we perform well against previous methods, achieving new state-of-the-art results on the challenging Meta-Dataset.

\textbf{Few-shot domain generalization.}
We also evaluate our method on few-shot domain generalization~\cite{du2020metanorm}, which combines the challenges of both few-shot classification and domain generalization. For few-shot domain generalization, each task has only a few samples in the support set for training and we test the model on tasks in a query set, which come from a different domain than the support set. 
The results are reported in Table~\ref{tab:fewdg}.  MetaKernel obtains the best performance, surpassing the MetaNorm~\cite{du2020metanorm} by a margin of up to $2.0\%$ on the $5$-way $1$-shot and $1.8\%$ on the $5$-way $5$-shot setting. Its performance on the few-shot domain generalization task demonstrates that MetaKernel  is not only able to handle the problem of few-shot learning, but also thrives under domain-shifts.

\subsection{Few-Shot Regression}
We also consider  regression tasks with a varying number of shots $k$, and compare MetaKernel with MAML~\cite{finn2017model}, a representative meta-learning algorithm. We follow  MAML  \cite{finn2017model} and fit a target sine function $y{=}A \sin{(wx + b)}$, with only a few annotated samples. $A \in [0.1, 5]$, $w \in [0.8, 1.2]$, and $ b\in [0, \pi ]$ denote the amplitude, frequency, and phase, which follow a uniform distribution within the corresponding interval. The goal is to estimate the target sine function given only $n$ randomly sampled data points. Here, we consider inputs within the range of $x\in [-5, 5]$, and conduct three tests under the conditions of $k {=} 3, 5, 10$. For fair comparison, we compute the feature embedding using a small MLP  with two hidden layers of size $40$, following the same settings used in MAML. 

The results in Figure~\ref{fig:reg} show that MetaKernel fits the function well with only three shots, even when we do not use the full model. It performs better with an increasing number of shots, almost entirely fitting the target function with ten shots. We observe all MetaKernel variants perform  better than MAML~\cite{finn2017model} for all three settings with varying numbers of shots, both visually and in terms of MSE. Best results are obtained with our full model.

\begin{figure*}[t]
	\centering
	\begin{minipage}{.48\textwidth}

	\begin{subfigure}
		\centering
		\includegraphics[width=0.482\columnwidth]{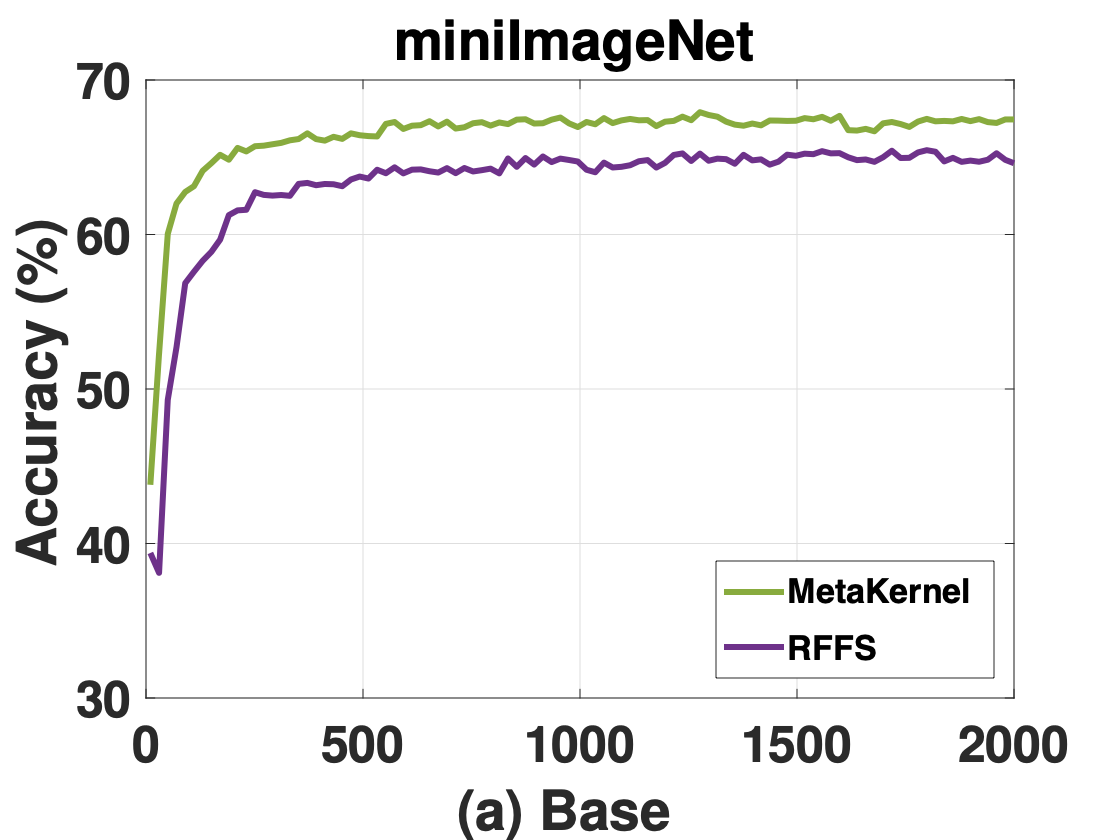}
		\label{fig:eff1}
	\end{subfigure}%
	\begin{subfigure}
		\centering
		\includegraphics[width=0.482\columnwidth]{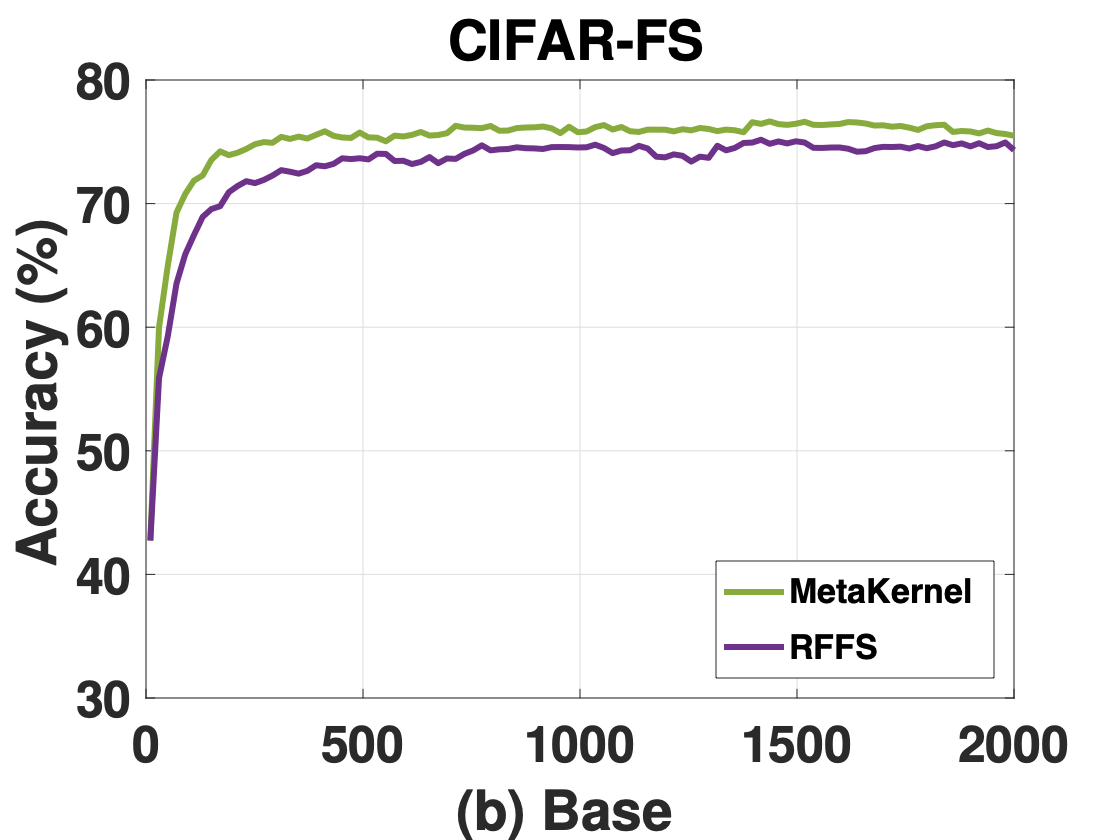}
		\label{fig:eff2}
	\end{subfigure}
	\vspace{-4mm}
	\caption{Efficiency with varying numbers $D$ of bases. MetaKernel consistently achieves better performance than regular RFFs, especially with relatively low sampling rates.}
	\label{fig:eff} 

	\end{minipage}
	\hspace{4mm}
	\begin{minipage}{.48\textwidth}
	\centering
	\vspace{-6mm}
	\begin{subfigure}
		\centering
		\includegraphics[width=0.482\columnwidth]{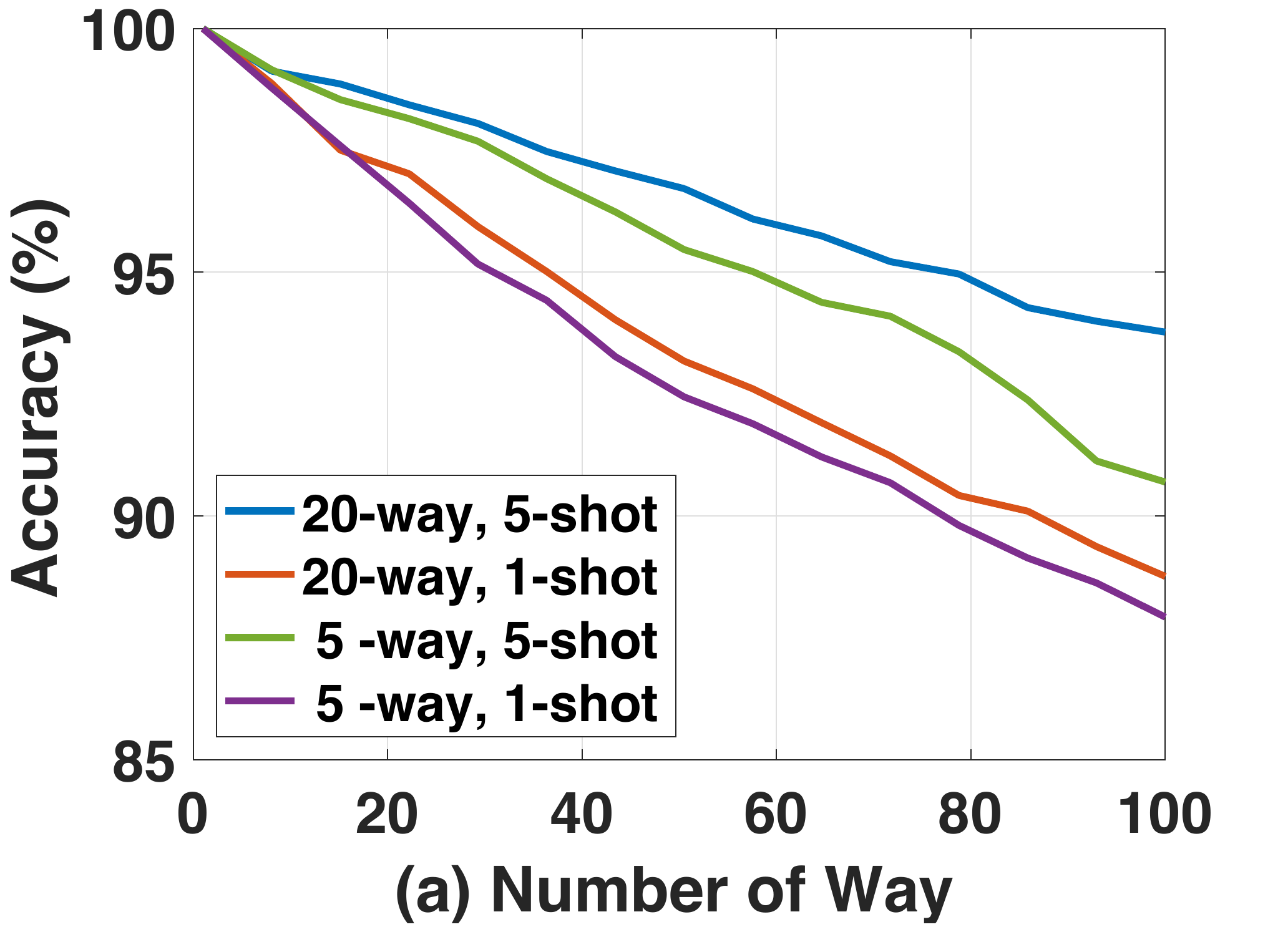}
	\end{subfigure}%
	\begin{subfigure}
		\centering
		\includegraphics[width=0.482\columnwidth]{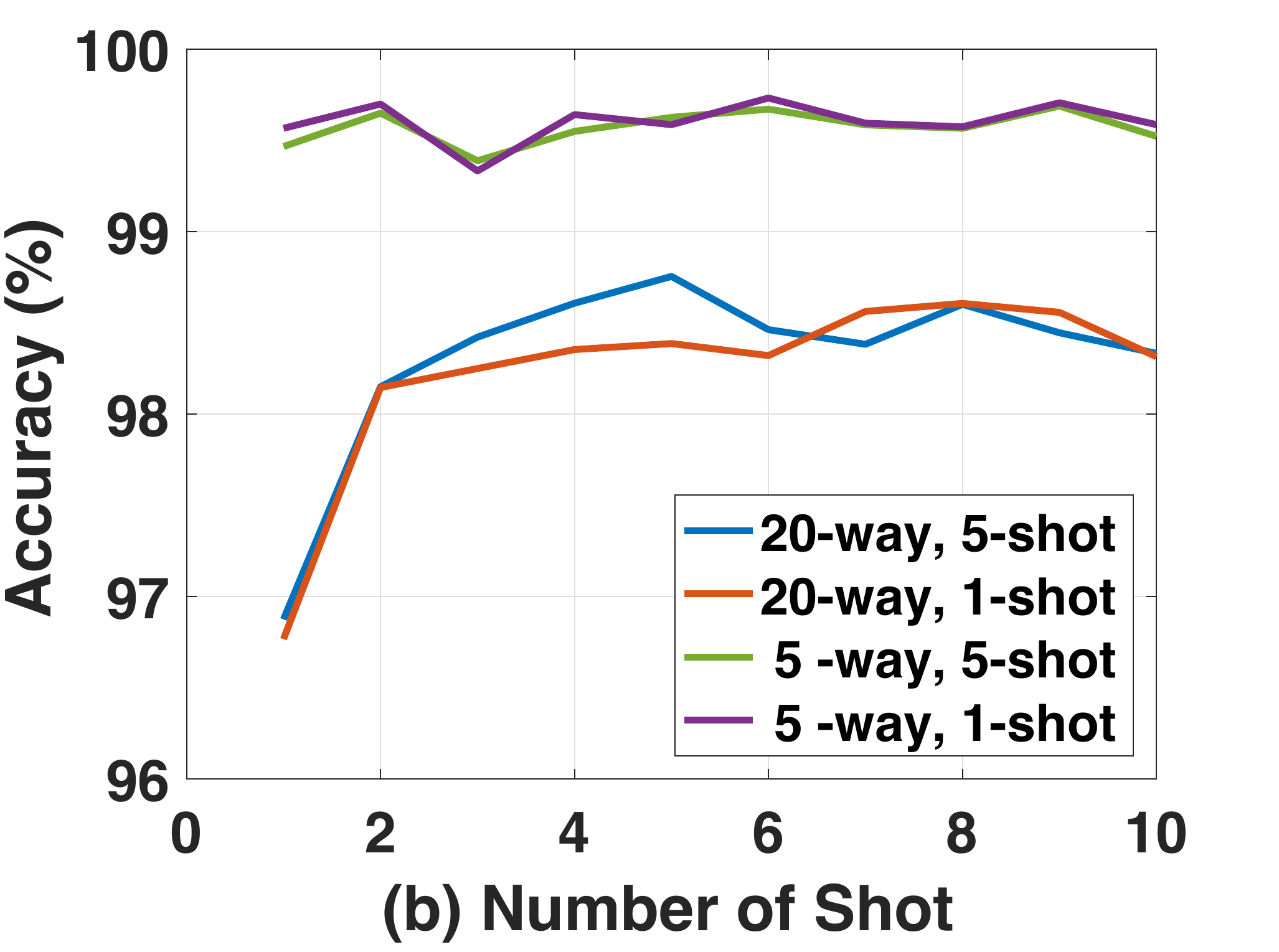}
	\end{subfigure}
	\caption{Versatility of MetaKernel with varied ways and shots on Omniglot.}
	\label{fig:flex}
		\end{minipage}
\end{figure*}

\subsection{Ablation Studies}
To study how our proposed components bring performance gains to MetaKernel on few-shot learning, our ablations consider: (1) the benefit of random Fourier features; (2) the benefit of task context inference; (3) the benefit of enriching random features by normalizing flows; (4) the effect of deeper embeddings; (5) the efficiency of the model; (6) the versatility of the model.


\textbf{Benefit of random Fourier features.} We first show the benefit of random Fourier features (RFFs) by comparing them with the regular RBF kernel. As can be seen from the first two rows in Table~\ref{tab:mini_kernel}, RFFs perform 10.7\% better than an RBF kernel on the $5$-way $1$-shot setting of \mini{}, and 14.9\% better on the $5$-way $5$-shot setting of \cifarfs{}. The considerable performance gain over RBF kernels on both datasets indicates the benefit of adaptive kernels based on random Fourier features for few-shot image classification. The modest performance obtained by RBF kernels is due to the mean of pair-wise distances of support samples being unable to provide a proper estimate of the kernel bandwidth. Note that the performance of RFFs is better than the variational RFFs on the $5$-way $1$-shot setting of \mini{}. This may be due to the fact that the support samples are too small, resulting in the random bases generated from the samples not accurately representing the current task, while the parameters in the random bases of RFFs are sampled from a standard Gaussian distribution.  Therefore, the context information among previous related tasks should be integrated into the variational RFFs. In addition, RFFs cannot use the context information directly since it generates random base parameters sampled from a deterministic distribution. 

\textbf{Benefit of task context inference.}
We investigate the benefit of adding task context inference to the MetaKernel. Specifically, we leverage a bi-\lstm~cell state $\textbf{c}$ to store and accrue the meta-knowledge shared among related tasks. The experimental results are reported in Table~\ref{tab:mini_kernel}. Adding task context inference on top of the MetaKernel with variational random features leads to a consistent gain under all settings, for both datasets. This demonstrates the effectiveness of using an \lstm~to explore task dependency.  

\textbf{Benefit of enriching features by normalizing flows.}
We show the benefit of enriching the variational random features by conditional normalizing flow in the last row of Table~\ref{tab:mini_kernel}.
we  find that MetaKernel performs better than MetaVRF (55.5\% -up 1.3\%) under the $5$-way $1$-shot setting on \mini{} and (64.3\% -up 1.2\%) under the $5$-way $1$-shot setting on \cifarfs{}.
These results indicate that the CNFs  provide more informative kernels for the new task, which allows the learned distribution of random bases to more closely approximate the real random bases distribution and therefore improves few-shot classification performance. 


\renewcommand{\arraystretch}{1.3}
\begin{table}[t]
\caption{Deep embedding performance ($\%$) on \mini{} (5-way) using a 28-layer wide residual feature extractor. The best performing methods and any other runs within the 95\% confidence margin are denoted in bold. MetaKernel also outperforms previous methods for deeper networks. }\label{tab:mini}
\centering
\begin{tabular}{lcc}
\toprule
\textbf{Method} & 1-shot & 5-shot \\
\midrule
\textbf{\textsc{Meta-SGD}}~\cite{li2017meta} & 54.24$\pm$0.03& 70.86$\pm$0.04  \\
\textbf{\textsc{SNAIL}}~\cite{mishra2018simple} & 55.71$\pm$0.99& 68.88$\pm$0.92  \\
\textbf{\textsc{Gidaris \etal}}~\cite{gidaris2018dynamic}& 56.20$\pm$0.86& 73.00$\pm$0.64  \\
\textbf{\textsc{Bauer \etal}}~\cite{bauer2017discriminative}& 56.30$\pm$0.40& 73.90$\pm$0.30  \\
\textbf{\textsc{adaResNet (DF)}}~\cite{munkhdalai2017rapid}& 57.10$\pm$0.70& 70.04$\pm$0.63  \\
\textbf{\textsc{TADAM}}~\cite{oreshkin2018tadam} & 58.50$\pm$0.30& 76.70$\pm$0.30  \\
\textbf{\textsc{Qiao \etal}}~\cite{qiao2018few}& 59.60$\pm$0.41& 73.54$\pm$0.19  \\
\textbf{\textsc{LEO}}~\cite{rusu2018meta} & 61.76$\pm$0.08& 77.59$\pm$0.12  \\
\textbf{\textsc{MetaVRF}}~\cite{zhen2020learning} & 63.80$\pm$0.05  & 77.97$\pm$0.28 \\
\rowcolor{Gray}
\textbf{\textsc{MetaKernel}} & \textbf{65.03}$\pm$0.03  & \textbf{79.01}$\pm$0.28 \\
\bottomrule
\end{tabular}
\end{table}

\textbf{Deep embeddings.} MetaKernel is independent of the convolutional architecture for feature extraction and works with deeper embeddings, either pre-trained or trained from scratch. In general, the performance improves with more powerful feature extraction architectures. We evaluate our method using pre-trained embeddings in order to compare with existing methods using deep embedding architectures.
Specifically, we adopt the pre-trained embeddings from a 28-layer wide residual network (WRN-28-10) \cite{zagoruyko2016wide}, in a similar fashion to \cite{rusu2018meta, bauer2017discriminative, qiao2018few}. We choose activations in the 21-st layer, with average pooling over spatial dimensions, as feature embeddings. The dimension of the pre-trained embeddings is $640$. We show the comparison results on the \mini~dataset for 5-way 1-shot and 5-shot settings in Table~\ref{tab:mini}. MetaKernel achieves the best performance under both settings and surpasses LEO~\cite{rusu2018meta}, a recently proposed meta-learning method, especially on the challenging 5-way 1-shot setting. Compared with our conference paper, MetaVRF~\cite{zhen2020learning}, MetaKernel performs 1.23\% better on the $5$-way $1$-shot setting of \mini{}, which also validates the effectiveness of the CNFs. The consistent state-of-the-art results on all benchmarks using both shallow and deep feature extraction networks validate the effectiveness of MetaKernel for few-shot learning.

\textbf{Efficiency.} Regular RFFs usually require high sampling rates to achieve satisfactory performance. However, our MetaKernel achieves high performance with a relatively low sampling rate, which guarantees its high efficiency. In Figure~\ref{fig:eff}, we compare with regular RFFs using different sampling rates. We provide the performance change of fully trained models using RFFs and MetaKernel under a varying number of bases $D$. We show the comparison results for the $5$-way $5$-shot setting on \mini~ and CIFAR-FS in Figure~\ref{fig:eff}. MetaKernel consistently yields higher performance than regular RFFs with the same number of sampled bases. The results verify the efficiency of MetaKernel in learning adaptive kernels and its effectiveness in improving performance by exploring the dependencies of related tasks.

\textbf{Versatility.} In contrast to most existing meta-learning methods, MetaKernel is applicable to versatile settings.
We evaluate the performance of MetaKernel on more challenging scenarios where the number of ways $N$ and  shots $k$ between training and testing are inconsistent. Specifically, we test the performance of MetaKernel on Omniglot tasks with varied $N$ and $k$, when it is trained on one particular $N$-way $k$-shot task. As shown in Figure~\ref{fig:flex}, the results demonstrate the trained model still produces good performance, even under the challenging conditions with a far higher number of ways. In particular, the model trained on the $20$-way $5$-shot task retains a high accuracy of $94\%$ on the $100$-way setting, as shown in Figure~\ref{fig:flex}(a). The results also indicate that our model exhibits considerable robustness and flexibility to a variety of testing conditions.

\section{Conclusion}
\label{sec:conclusion}
In this paper, we introduce kernel approximation based on random Fourier features into the meta-learning framework for few-shot learning. We propose to learn random features for each few-shot task in a data-driven way by formulating it as a variational inference problem, where the random Fourier basis is defined as the latent variable. We introduce an inference network based on an LSTM module, which enables the shared knowledge from related tasks to be incorporated into each individual task. To further enhance the kernels, we introduce conditional normalizing flows to generate richer posteriors over random bases, resulting in more informative random features. Experimental results on both regression and classification tasks demonstrate the effectiveness for few-shot learning. The extensive ablation study demonstrates the efficacy of each component in our MetaKernel.

\bibliographystyle{IEEEtran}
\bibliography{MetaKernel}

\begin{thebibliography}{100}
\providecommand{\url}[1]{#1}
\csname url@samestyle\endcsname
\providecommand{\newblock}{\relax}
\providecommand{\bibinfo}[2]{#2}
\providecommand{\BIBentrySTDinterwordspacing}{\spaceskip=0pt\relax}
\providecommand{\BIBentryALTinterwordstretchfactor}{4}
\providecommand{\BIBentryALTinterwordspacing}{\spaceskip=\fontdimen2\font plus
\BIBentryALTinterwordstretchfactor\fontdimen3\font minus
  \fontdimen4\font\relax}
\providecommand{\BIBforeignlanguage}[2]{{%
\expandafter\ifx\csname l@#1\endcsname\relax
\typeout{** WARNING: IEEEtran.bst: No hyphenation pattern has been}%
\typeout{** loaded for the language `#1'. Using the pattern for}%
\typeout{** the default language instead.}%
\else
\language=\csname l@#1\endcsname
\fi
#2}}
\providecommand{\BIBdecl}{\relax}
\BIBdecl

\bibitem{zhen2020learning}
X.~Zhen, H.~Sun, Y.~Du, J.~Xu, Y.~Yin, L.~Shao, and C.~G.~M. Snoek, ``Learning
  to learn kernels with variational random features,'' in \emph{International
  Conference on Machine Learning}.\hskip 1em plus 0.5em minus 0.4em\relax PMLR,
  2020, pp. 11\,409--11\,419.

\bibitem{krizhevsky2012imagenet}
A.~Krizhevsky, I.~Sutskever, and G.~E. Hinton, ``{ImageNet} classification with
  deep convolutional neural networks,'' in \emph{Advances in Neural Information
  Processing Systems}, 2012, pp. 1097--1105.

\bibitem{resnet}
K.~He, X.~Zhang, S.~Ren, and J.~Sun, ``Deep residual learning for image
  recognition,'' in \emph{IEEE Conference on Computer Vision and Pattern
  Recognition}, 2016, pp. 770--778.

\bibitem{imagenet_cvpr09}
J.~Deng, W.~Dong, R.~Socher, L.-J. Li, K.~Li, and L.~Fei-Fei, ``{ImageNet: A
  Large-Scale Hierarchical Image Database},'' in \emph{IEEE Conference on
  Computer Vision and Pattern Recognition}, 2009.

\bibitem{fei2006one}
L.~Fei-Fei, R.~Fergus, and P.~Perona, ``One-shot learning of object
  categories,'' \emph{IEEE Transactions on Pattern Analysis and Machine
  Intelligence}, vol.~28, no.~4, pp. 594--611, 2006.

\bibitem{lake2015human}
B.~M. Lake, R.~Salakhutdinov, and J.~B. Tenenbaum, ``Human-level concept
  learning through probabilistic program induction,'' \emph{Science}, vol. 350,
  no. 6266, pp. 1332--1338, 2015.

\bibitem{ravi2017optimization}
S.~Ravi and H.~Larochelle, ``Optimization as a model for few-shot learning,''
  in \emph{International Conference on Learning Representations}, 2017.

\bibitem{finn2017model}
C.~Finn, P.~Abbeel, and S.~Levine, ``Model-agnostic meta-learning for fast
  adaptation of deep networks,'' in \emph{International Conference on Machine
  Learning}.\hskip 1em plus 0.5em minus 0.4em\relax JMLR. org, 2017, pp.
  1126--1135.

\bibitem{Schmidhuber1992}
J.~{Schmidhuber}, ``Learning to control fast-weight memories: An alternative to
  dynamic recurrent networks,'' \emph{Neural Computation}, vol.~4, no.~1, pp.
  131--139, 1992.

\bibitem{thrun2012learning}
S.~Thrun and L.~Pratt, \emph{Learning to learn}.\hskip 1em plus 0.5em minus
  0.4em\relax Springer Science \& Business Media, 2012.

\bibitem{andrychowicz2016learning}
M.~Andrychowicz, M.~Denil, S.~Gomez, M.~W. Hoffman, D.~Pfau, T.~Schaul,
  B.~Shillingford, and N.~de~Freitas, ``Learning to learn by gradient descent
  by gradient descent,'' in \emph{Advances in Neural Information Processing
  Systems}, 2016.

\bibitem{rusu2018meta}
A.~A. Rusu, D.~Rao, J.~Sygnowski, O.~Vinyals, R.~Pascanu, S.~Osindero, and
  R.~Hadsell, ``Meta-learning with latent embedding optimization,'' in
  \emph{International Conference on Learning Representations}, 2019.

\bibitem{vinyals2016matching}
O.~Vinyals, C.~Blundell, T.~Lillicrap, D.~Wierstra \emph{et~al.}, ``Matching
  networks for one shot learning,'' in \emph{Advances in Neural Information
  Processing Systems}, 2016, pp. 3630--3638.

\bibitem{snell2017prototypical}
J.~Snell, K.~Swersky, and R.~Zemel, ``Prototypical networks for few-shot
  learning,'' in \emph{Advances in Neural Information Processing Systems},
  2017, pp. 4077--4087.

\bibitem{gordon2018meta}
J.~Gordon, J.~Bronskill, M.~Bauer, S.~Nowozin, and R.~E. Turner,
  ``Meta-learning probabilistic inference for prediction,'' in
  \emph{International Conference on Learning Representations}, 2019.

\bibitem{du2020metanorm}
Y.~Du, X.~Zhen, L.~Shao, and C.~G.~M. Snoek, ``{MetaNorm}: Learning to
  normalize few-shot batches across domains,'' in \emph{International
  Conference on Learning Representations}, 2021.

\bibitem{smola1998learning}
B.~Sch{\"o}lkopf and A.~J. Smola, \emph{Learning with kernels}.\hskip 1em plus
  0.5em minus 0.4em\relax MIT Press, 2002.

\bibitem{scholkopf2018learning}
B.~Scholkopf and A.~J. Smola, \emph{Learning with kernels: support vector
  machines, regularization, optimization, and beyond}.\hskip 1em plus 0.5em
  minus 0.4em\relax Adaptive Computation and Machine Learning series, 2018.

\bibitem{hofmann2008kernel}
T.~Hofmann, B.~Sch{\"o}lkopf, and A.~J. Smola, ``Kernel methods in machine
  learning,'' \emph{The Annals of Statistics}, pp. 1171--1220, 2008.

\bibitem{cristianini2000introduction}
N.~Cristianini, J.~Shawe-Taylor \emph{et~al.}, \emph{An introduction to support
  vector machines and other kernel-based learning methods}.\hskip 1em plus
  0.5em minus 0.4em\relax Cambridge University Press, 2000.

\bibitem{smola2004tutorial}
A.~J. Smola and B.~Sch{\"o}lkopf, ``A tutorial on support vector regression,''
  \emph{Statistics and Computing}, 2004.

\bibitem{rahimi2007random}
A.~Rahimi and B.~Recht, ``Random features for large-scale kernel machines,'' in
  \emph{Advances in Neural Information Processing Systems}, 2007.

\bibitem{sinha2016learning}
A.~Sinha and J.~C. Duchi, ``Learning kernels with random features.'' in
  \emph{Advances in Neural Information Processing Systems}, 2016.

\bibitem{bach2004multiple}
F.~R. Bach, G.~R. Lanckriet, and M.~I. Jordan, ``Multiple kernel learning,
  conic duality, and the smo algorithm,'' in \emph{International Conference on
  Machine Learning}, 2004, p.~6.

\bibitem{hensman2017variational}
J.~Hensman, N.~Durrande, and A.~Solin, ``Variational fourier features for
  gaussian processes,'' \emph{Journal of Machine Learning Research}, vol.~18,
  no.~1, pp. 5537--5588, 2017.

\bibitem{carratino2018learning}
L.~Carratino, A.~Rudi, and L.~Rosasco, ``Learning with sgd and random
  features,'' in \emph{Advances in Neural Information Processing Systems},
  2018, pp. 10\,192--10\,203.

\bibitem{bullins2018not}
B.~Bullins, C.~Zhang, and Y.~Zhang, ``Not-so-random features,'' in
  \emph{International Conference on Learning Representations}, 2018.

\bibitem{li2019implicit}
C.-L. Li, W.-C. Chang, Y.~Mroueh, Y.~Yang, and B.~Poczos, ``Implicit kernel
  learning,'' in \emph{International Conference on Artificial Intelligence and
  Statistics}, 2019, pp. 2007--2016.

\bibitem{hochreiter1997long}
S.~Hochreiter and J.~Schmidhuber, ``Long short-term memory,'' \emph{Neural
  computation}, vol.~9, no.~8, pp. 1735--1780, 1997.

\bibitem{dinh2014nice}
L.~Dinh, D.~Krueger, and Y.~Bengio, ``Nice: Non-linear independent components
  estimation,'' \emph{arXiv preprint arXiv:1410.8516}, 2014.

\bibitem{dinh2016density}
L.~Dinh, J.~Sohl-Dickstein, and S.~Bengio, ``Density estimation using real
  nvp,'' \emph{International Conference on Learning Representations}, 2017.

\bibitem{rezende2015variational}
D.~Rezende and S.~Mohamed, ``Variational inference with normalizing flows,'' in
  \emph{International Conference on Machine Learning}.\hskip 1em plus 0.5em
  minus 0.4em\relax PMLR, 2015, pp. 1530--1538.

\bibitem{kingma2018glow}
D.~P. Kingma and P.~Dhariwal, ``Glow: generative flow with invertible 1$\times$
  1 convolutions,'' in \emph{Advances in Neural Information Processing
  Systems}, 2018, pp. 10\,236--10\,245.

\bibitem{winkler2019learning}
C.~Winkler, D.~Worrall, E.~Hoogeboom, and M.~Welling, ``Learning likelihoods
  with conditional normalizing flows,'' \emph{arXiv preprint arXiv:1912.00042},
  2019.

\bibitem{triantafillou2019meta}
E.~Triantafillou, T.~Zhu, V.~Dumoulin, P.~Lamblin, U.~Evci, K.~Xu, R.~Goroshin,
  C.~Gelada, K.~Swersky, P.-A. Manzagol, and H.~Larochelle, ``Meta-dataset: A
  dataset of datasets for learning to learn from few examples,'' \emph{arXiv
  preprint arXiv:1903.03096}, 2019.

\bibitem{rajeswaran2019meta}
A.~Rajeswaran, C.~Finn, S.~Kakade, and S.~Levine, ``Meta-learning with implicit
  gradients,'' \emph{arXiv preprint arXiv:1909.04630}, 2019.

\bibitem{hospedales2020meta}
T.~Hospedales, A.~Antoniou, P.~Micaelli, and A.~Storkey, ``Meta-learning in
  neural networks: A survey,'' \emph{arXiv preprint arXiv:2004.05439}, 2020.

\bibitem{allen2019infinite}
K.~R. Allen, E.~Shelhamer, H.~Shin, and J.~B. Tenenbaum, ``Infinite mixture
  prototypes for few-shot learning,'' in \emph{International Conference on
  Machine Learning}, 2019, pp. 232--241.

\bibitem{oreshkin2018tadam}
B.~Oreshkin, P.~R. L{\'o}pez, and A.~Lacoste, ``Tadam: Task dependent adaptive
  metric for improved few-shot learning,'' in \emph{Advances in Neural
  Information Processing Systems}, 2018, pp. 721--731.

\bibitem{yoon2019tapnet}
S.~W. Yoon, J.~Seo, and J.~Moon, ``Tapnet: Neural network augmented with
  task-adaptive projection for few-shot learning,'' in \emph{International
  Conference on Machine Learning}.\hskip 1em plus 0.5em minus 0.4em\relax PMLR,
  2019, pp. 7115--7123.

\bibitem{garcia2018few}
V.~Garcia and J.~Bruna, ``Few-shot learning with graph neural networks,'' in
  \emph{International Conference on Learning Representations}, 2018.

\bibitem{cao2019theoretical}
T.~Cao, M.~Law, and S.~Fidler, ``A theoretical analysis of the number of shots
  in few-shot learning,'' \emph{arXiv preprint arXiv:1909.11722}, 2019.

\bibitem{zhen2020memory}
X.~Zhen, Y.~Du, H.~Xiong, Q.~Qiu, C.~G.~M. Snoek, and L.~Shao, ``Learning to
  learn variational semantic memory,'' in \emph{Advances in Neural Information
  Processing Systems}, 2020.

\bibitem{zintgraf2019fast}
L.~Zintgraf, K.~Shiarli, V.~Kurin, K.~Hofmann, and S.~Whiteson, ``Fast context
  adaptation via meta-learning,'' in \emph{International Conference on Machine
  Learning}, 2019, pp. 7693--7702.

\bibitem{chen2017learning}
Y.~Chen, M.~W. Hoffman, S.~G. Colmenarejo, M.~Denil, T.~P. Lillicrap,
  M.~Botvinick, and N.~De~Freitas, ``Learning to learn without gradient descent
  by gradient descent,'' in \emph{International Conference on Machine
  Learning}, 2017, pp. 748--756.

\bibitem{edwards2016towards}
H.~Edwards and A.~Storkey, ``Towards a neural statistician,'' \emph{arXiv
  preprint arXiv:1606.02185}, 2016.

\bibitem{finn2018probabilistic}
C.~Finn, K.~Xu, and S.~Levine, ``Probabilistic model-agnostic meta-learning,''
  in \emph{Advances in Neural Information Processing Systems}, 2018, pp.
  9516--9527.

\bibitem{saemundsson2018meta}
S.~S{\ae}mundsson, K.~Hofmann, and M.~P. Deisenroth, ``Meta reinforcement
  learning with latent variable gaussian processes,'' \emph{arXiv preprint
  arXiv:1803.07551}, 2018.

\bibitem{bertinetto2018meta}
L.~Bertinetto, J.~F. Henriques, P.~H. Torr, and A.~Vedaldi, ``Meta-learning
  with differentiable closed-form solvers,'' in \emph{International Conference
  on Learning Representations}, 2019.

\bibitem{santoro2016meta}
A.~Santoro, S.~Bartunov, M.~Botvinick, D.~Wierstra, and T.~Lillicrap,
  ``Meta-learning with memory-augmented neural networks,'' in
  \emph{International Conference on Machine Learning}, 2016, pp. 1842--1850.

\bibitem{munkhdalai2017meta}
T.~Munkhdalai and H.~Yu, ``Meta networks,'' in \emph{International Conference
  on Machine Learning}, 2017.

\bibitem{munkhdalai2017rapid}
T.~Munkhdalai, X.~Yuan, S.~Mehri, and A.~Trischler, ``Rapid adaptation with
  conditionally shifted neurons,'' \emph{arXiv preprint arXiv:1712.09926},
  2017.

\bibitem{bishop2006pattern}
C.~M. Bishop, \emph{Pattern Recognition and Machine Learning}.\hskip 1em plus
  0.5em minus 0.4em\relax springer, 2006.

\bibitem{shervashidze2011weisfeiler}
N.~Shervashidze, P.~Schweitzer, E.~J.~v. Leeuwen, K.~Mehlhorn, and K.~M.
  Borgwardt, ``Weisfeiler-lehman graph kernels,'' \emph{Journal of Machine
  Learning Research}, vol.~12, no. Sep, pp. 2539--2561, 2011.

\bibitem{gonen2011multiple}
M.~G{\"o}nen and E.~Alpayd{\i}n, ``Multiple kernel learning algorithms,''
  \emph{Journal of Machine Learning Research}, vol.~12, pp. 2211--2268, 2011.

\bibitem{duvenaud2013structure}
D.~Duvenaud, J.~R. Lloyd, R.~Grosse, J.~B. Tenenbaum, and Z.~Ghahramani,
  ``Structure discovery in nonparametric regression through compositional
  kernel search,'' \emph{arXiv preprint arXiv:1302.4922}, 2013.

\bibitem{rahimi2008random}
A.~Rahimi and B.~Recht, ``Random features for large-scale kernel machines,'' in
  \emph{Advances in Neural Information Processing Systems}, 2007, pp.
  1177--1184.

\bibitem{gartner2002multi}
T.~G{\"a}rtner, P.~A. Flach, A.~Kowalczyk, and A.~J. Smola, ``Multi-instance
  kernels,'' in \emph{International Conference on Machine Learning}, 2002.

\bibitem{rudin1962fourier}
W.~Rudin, \emph{Fourier analysis on groups}.\hskip 1em plus 0.5em minus
  0.4em\relax Wiley Online Library, 1962, vol. 121967.

\bibitem{wilson2013gaussian}
A.~Wilson and R.~Adams, ``Gaussian process kernels for pattern discovery and
  extrapolation,'' in \emph{International Conference on Machine Learning},
  2013, pp. 1067--1075.

\bibitem{yang2015carte}
Z.~Yang, A.~Wilson, A.~Smola, and L.~Song, ``A la carte--learning fast
  kernels,'' in \emph{Artificial Intelligence and Statistics}, 2015, pp.
  1098--1106.

\bibitem{avron2016quasi}
H.~Avron, V.~Sindhwani, J.~Yang, and M.~W. Mahoney, ``Quasi-monte carlo feature
  maps for shift-invariant kernels,'' \emph{Journal of Machine Learning
  Research}, vol.~17, no.~1, pp. 4096--4133, 2016.

\bibitem{chang2017data}
W.-C. Chang, C.-L. Li, Y.~Yang, and B.~Poczos, ``Data-driven random fourier
  features using stein effect,'' \emph{arXiv preprint arXiv:1705.08525}, 2017.

\bibitem{papamakarios2021normalizing}
G.~Papamakarios, E.~Nalisnick, D.~J. Rezende, S.~Mohamed, and
  B.~Lakshminarayanan, ``Normalizing flows for probabilistic modeling and
  inference,'' \emph{Journal of Machine Learning Research}, vol.~22, no.~57,
  pp. 1--64, 2021.

\bibitem{kingma2016improved}
D.~P. Kingma, T.~Salimans, R.~Jozefowicz, X.~Chen, I.~Sutskever, and
  M.~Welling, ``Improved variational inference with inverse autoregressive
  flow,'' \emph{Advances in Neural Information Processing Systems}, vol.~29,
  pp. 4743--4751, 2016.

\bibitem{papamakarios2017masked}
G.~Papamakarios, T.~Pavlakou, and I.~Murray, ``Masked autoregressive flow for
  density estimation,'' in \emph{Advances in Neural Information Processing
  Systems}, 2017, pp. 2335--2344.

\bibitem{chen2019residual}
R.~T. Chen, J.~Behrmann, D.~Duvenaud, and J.-H. Jacobsen, ``Residual flows for
  invertible generative modeling,'' in \emph{Advances in Neural Information
  Processing Systems}, 2019.

\bibitem{ho2019flow++}
J.~Ho, X.~Chen, A.~Srinivas, Y.~Duan, and P.~Abbeel, ``Flow++: Improving
  flow-based generative models with variational dequantization and architecture
  design,'' in \emph{International Conference on Machine Learning}.\hskip 1em
  plus 0.5em minus 0.4em\relax PMLR, 2019, pp. 2722--2730.

\bibitem{esling2019universal}
P.~Esling, N.~Masuda, A.~Bardet, R.~Despres \emph{et~al.}, ``Universal audio
  synthesizer control with normalizing flows,'' \emph{arXiv preprint
  arXiv:1907.00971}, 2019.

\bibitem{prenger2019waveglow}
R.~Prenger, R.~Valle, and B.~Catanzaro, ``Waveglow: A flow-based generative
  network for speech synthesis,'' in \emph{IEEE International Conference on
  Acoustics, Speech and Signal Processing}, 2019, pp. 3617--3621.

\bibitem{jacobsen2018revnet}
J.-H. Jacobsen, A.~Smeulders, and E.~Oyallon, ``i-revnet: Deep invertible
  networks,'' in \emph{International Conference on Learning Representations},
  2018.

\bibitem{sohn2015learning}
K.~Sohn, H.~Lee, and X.~Yan, ``Learning structured output representation using
  deep conditional generative models,'' in \emph{Advances in Neural Information
  Processing Systems}, 2015, pp. 3483--3491.

\bibitem{kingma2013auto}
D.~P. Kingma and M.~Welling, ``Auto-encoding variational bayes,'' \emph{arXiv
  preprint arXiv:1312.6114}, 2013.

\bibitem{rezende2014stochastic}
D.~J. Rezende, S.~Mohamed, and D.~Wierstra, ``Stochastic backpropagation and
  approximate inference in deep generative models,'' \emph{arXiv preprint
  arXiv:1401.4082}, 2014.

\bibitem{gers2000recurrent}
F.~A. Gers and J.~Schmidhuber, ``Recurrent nets that time and count,'' in
  \emph{Proceedings of the IEEE-INNS-ENNS International Joint Conference on
  Neural Networks}, vol.~3, 2000, pp. 189--194.

\bibitem{schuster1997bidirectional}
M.~Schuster and K.~K. Paliwal, ``Bidirectional recurrent neural networks,''
  \emph{IEEE Transactions on Signal Processing}, vol.~45, no.~11, pp.
  2673--2681, 1997.

\bibitem{graves2005framewise}
A.~Graves and J.~Schmidhuber, ``Framewise phoneme classification with
  bidirectional lstm and other neural network architectures,'' \emph{Neural
  Networks}, vol.~18, no. 5-6, pp. 602--610, 2005.

\bibitem{zaheer2017deep}
M.~Zaheer, S.~Kottur, S.~Ravanbakhsh, B.~Poczos, R.~R. Salakhutdinov, and A.~J.
  Smola, ``Deep sets,'' in \emph{Advances in Neural Information Processing
  Systems}, 2017, pp. 3391--3401.

\bibitem{kim2019attentive}
H.~Kim, A.~Mnih, J.~Schwarz, M.~Garnelo, A.~Eslami, D.~Rosenbaum, O.~Vinyals,
  and Y.~W. Teh, ``Attentive neural processes,'' in \emph{International
  Conference on Learning Representations}, 2019.

\bibitem{krizhevsky2009learning}
A.~Krizhevsky, ``Learning multiple layers of features from tiny images,''
  University of Toronto, Tech. Rep., 2009.

\bibitem{peng2019moment}
X.~Peng, Q.~Bai, X.~Xia, Z.~Huang, K.~Saenko, and B.~Wang, ``Moment matching
  for multi-source domain adaptation,'' in \emph{IEEE International Conference
  on Computer Vision}, 2019, pp. 1406--1415.

\bibitem{russakovsky2015imagenet}
O.~Russakovsky, J.~Deng, H.~Su, J.~Krause, S.~Satheesh, S.~Ma, Z.~Huang,
  A.~Karpathy, A.~Khosla, M.~Bernstein, A.~Berg, and L.~Fei-Fei, ``{ImageNet}
  large scale visual recognition challenge,'' \emph{International Journal of
  Computer Vision}, vol. 115, no.~3, pp. 211--252, 2015.

\bibitem{maji13finegrained}
S.~Maji, E.~Rahtu, J.~Kannala, M.~Blaschko, and A.~Vedaldi, ``Fine-grained
  visual classification of aircraft,'' \emph{arXiv preprint arXiv:1306.5151},
  2013.

\bibitem{Quick}
J.~Jongejan, H.~Rowley, T.~Kawashima, J.~Kim, and N.~Fox-Gieg, ``The quick,
  draw! – a.i. experiment,'' \url{quickdraw.withgoogle.com}, 2016.

\bibitem{Fungi}
B.~Schroeder and Y.~Cui, ``{FGVCx} fungi classification challenge 2018,''
  \url{github.com/ visipedia/fgvcx_fungi_comp}, 2018.

\bibitem{Houben-IJCNN-2013}
S.~Houben, J.~Stallkamp, J.~Salmen, M.~Schlipsing, and C.~Igel, ``Detection of
  traffic signs in real-world images: The {German} traffic sign detection
  benchmark,'' in \emph{International Joint Conference on Neural Networks},
  2013, pp. 1--8.

\bibitem{lin2014microsoft}
T.-Y. Lin, M.~Maire, S.~Belongie, J.~Hays, P.~Perona, D.~Ramanan,
  P.~Doll{\'a}r, and C.~L. Zitnick, ``Microsoft {COCO}: Common objects in
  context,'' in \emph{European Conference on Computer Vision}, 2014, pp.
  740--755.

\bibitem{WahCUB_200_2011}
C.~Wah, S.~Branson, P.~Welinder, P.~Perona, and S.~Belongie, ``The caltech-ucsd
  birds-200-2011 dataset,''
  \emph{http://www.vision.caltech.edu/visipedia/CUB-200-2011.html}, 2011.

\bibitem{cimpoi14describing}
M.~Cimpoi, S.~Maji, I.~Kokkinos, S.~Mohamed, and A.~Vedaldi, ``Describing
  textures in the wild,'' in \emph{{IEEE} Conference on Computer Vision and
  Pattern Recognition}, 2014.

\bibitem{Nilsback08}
M.-E. Nilsback and A.~Zisserman, ``Automated flower classification over a large
  number of classes,'' in \emph{Indian Conference on Computer Vision, Graphics
  \& Image Processing}, 2008, pp. 722--729.

\bibitem{snell2020bayesian}
J.~Snell and R.~Zemel, ``Bayesian few-shot classification with one-vs-each
  p\'olya-gamma augmented gaussian processes,'' \emph{arXiv preprint
  arXiv:2007.10417}, 2020.

\bibitem{sung2018learning}
F.~Sung, Y.~Yang, L.~Zhang, T.~Xiang, P.~H. Torr, and T.~M. Hospedales,
  ``Learning to compare: Relation network for few-shot learning,'' in
  \emph{IEEE Conference on Computer Vision and Pattern Recognition}, 2018, pp.
  1199--1208.

\bibitem{devosreproducing}
A.~Devos, S.~Chatel, and M.~Grossglauser, ``Reproducing meta-learning with
  differentiable closed-form solvers,'' in \emph{ICLR Workshop}, 2019.

\bibitem{tian2020rethinking}
Y.~Tian, Y.~Wang, D.~Krishnan, J.~B. Tenenbaum, and P.~Isola, ``Rethinking
  few-shot image classification: a good embedding is all you need?''
  \emph{arXiv preprint arXiv:2003.11539}, 2020.

\bibitem{yu2016orthogonal}
F.~X. Yu, A.~T. Suresh, K.~M. Choromanski, D.~N. Holtmann-Rice, and S.~Kumar,
  ``Orthogonal random features,'' in \emph{Advances in Neural Information
  Processing Systems}, 2016, pp. 1975--1983.

\bibitem{li2017meta}
Z.~Li, F.~Zhou, F.~Chen, and H.~Li, ``Meta-sgd: Learning to learn quickly for
  few-shot learning,'' \emph{arXiv preprint arXiv:1707.09835}, 2017.

\bibitem{mishra2018simple}
N.~Mishra, M.~Rohaninejad, X.~Chen, and P.~Abbeel, ``A simple neural attentive
  meta-learner,'' in \emph{International Conference on Learning
  Representations}, 2018.

\bibitem{gidaris2018dynamic}
S.~Gidaris and N.~Komodakis, ``Dynamic few-shot visual learning without
  forgetting,'' in \emph{IEEE Conference on Computer Vision and Pattern
  Recognition}, 2018, pp. 4367--4375.

\bibitem{bauer2017discriminative}
M.~Bauer, M.~Rojas-Carulla, J.~B. {\'S}wi{\k{a}}tkowski, B.~Sch{\"o}lkopf, and
  R.~E. Turner, ``Discriminative k-shot learning using probabilistic models,''
  \emph{arXiv preprint arXiv:1706.00326}, 2017.

\bibitem{qiao2018few}
S.~Qiao, C.~Liu, W.~Shen, and A.~L. Yuille, ``Few-shot image recognition by
  predicting parameters from activations,'' in \emph{IEEE Conference on
  Computer Vision and Pattern Recognition}, 2018, pp. 7229--7238.

\bibitem{zagoruyko2016wide}
S.~Zagoruyko and N.~Komodakis, ``Wide residual networks,'' \emph{arXiv preprint
  arXiv:1605.07146}, 2016.

\bibitem{kingma2014adam}
D.~P. Kingma and J.~Ba, ``Adam: A method for stochastic optimization,''
  \emph{arXiv preprint arXiv:1412.6980}, 2014.

\end{thebibliography}

\clearpage
\onecolumn

\section{Appendix}

\subsection{Derivations of the ELBO}
\label{ELBO}
For a singe task, we begin with maximizing log-likelihood of the conditional distribution $p(\mathbf{y} | \mathbf{x}, \mathcal{S})$ to derive the ELBO of MetaKernel. By leveraging Jensen's inequality, we have the following steps as
\begin{align}
\log  p(\mathbf{y} | \mathbf{x},\mathcal{S})&= \log  \int p(\mathbf{y} | \mathbf{x}, \mathcal{S}, \bm{\omega})  p(\bm{\omega} | \mathbf{x}, \mathcal{S}) d\bm{\omega} \\
&= \log  \int p(\mathbf{y} | \mathbf{x}, \mathcal{S}, \bm{\omega})  p(\bm{\omega} | \mathbf{x}, \mathcal{S}) \frac{q_{\phi}(\bm{\omega}| \mathcal{S})}{q_{\phi}(\bm{\omega}| \mathcal{S})} d\bm{\omega}\\
&\geq \int \log \left[ \frac{ p(\mathbf{y} | \mathbf{x}, \mathcal{S}, \bm{\omega})  p(\bm{\omega} | \mathbf{x}, \mathcal{S}) }{q_{\phi}(\bm{\omega}| \mathcal{S})} \right] q_{\phi}(\bm{\omega}|  \mathcal{S}) d\bm{\omega} \\
&= \underbrace{\mathbb{E}_{q_{\phi}(\bm{\omega}| \mathcal{S})} \log \, [p(\mathbf{y} | \mathbf{x},  \mathcal{S},\bm{\omega} )] - \KL[q_{\phi}(\bm{\omega}|\mathcal{S}) || p(\bm{\omega} | \mathbf{x}, \mathcal{S})]}_{\text{ELBO}}.
\label{der-likeli}
\end{align}

The ELBO  can also be derived from the perspective of the KL divergence between the variational posterior $q_{\phi}(\bm{\omega}| \mathcal{S})$ and the posterior $p(\bm{\omega} | \mathbf{y}, \mathbf{x}, \mathcal{S})$:
\begin{equation}
\begin{aligned}
\KL[q_{\phi}(\bm{\omega}| \mathcal{S}) || p(\bm{\omega} | \mathbf{y}, \mathbf{x}, \mathcal{S})]
& = \mathbb{E}_{q_{\phi}(\bm{\omega}| \mathcal{S})} \left[\log q_{\phi}(\bm{\omega}| \mathcal{S}) - \log p(\bm{\omega} | \mathbf{y}, \mathbf{x}, \mathcal{S})\right]\\
& = \mathbb{E}_{q_{\phi}(\bm{\omega}| \mathcal{S})} \left[\log q_{\phi}(\bm{\omega}| \mathcal{S}) - \log \frac{p(\mathbf{y} | \bm{\omega}, \mathbf{x}, \mathcal{S}) p(\bm{\omega}| \mathbf{x}, \mathcal{S})}{p(\mathbf{y}|  \mathbf{x}, \mathcal{S}) }\right] \\&= \log p(\mathbf{y}| \mathbf{x}, \mathcal{S}) + \mathbb{E}_{q_{\phi}(\bm{\omega}| \mathcal{S})} \left[\log q_{\phi}(\bm{\omega}| \mathcal{S})- \log p(\mathbf{y} | \bm{\omega}, \mathbf{x}, \mathcal{S}) - \log p(\bm{\omega}| \mathbf{x}, \mathcal{S})\right]\\
&= \log p(\mathbf{y}| \mathbf{x}, \mathcal{S}) - \mathbb{E}_{q_{\phi}(\bm{\omega}| \mathcal{S})} \left[ \log p(\mathbf{y} | \bm{\omega}, \mathbf{x}, \mathcal{S})\right] + \KL[ q_{\phi}(\bm{\omega}| \mathcal{S}) || p(\bm{\omega}| \mathbf{x}, \mathcal{S})] \geq 0.
\label{der-kl}
\end{aligned}
\end{equation}

Therefore, the lower bound of the $\log p(\mathbf{y}| \mathbf{x}, 
\mathcal{S})$ is 
\begin{equation}
\begin{aligned}
\log  p(\mathbf{y} | \mathbf{x}, \mathcal{S}) &\geq \mathbb{E}_{q_{\phi}(\bm{\omega}| \mathcal{S})} \log \, [p(\mathbf{y} | \mathbf{x}, \mathcal{S}, \bm{\omega} )]  - \KL[q_{\phi}(\bm{\omega}|\mathcal{S}) || p(\bm{\omega} | \mathbf{x}, \mathcal{S})],
\label{elbo}
\end{aligned}
\end{equation}
which is consistent with (\ref{der-likeli}).

\subsection{Cross attention in the prior network}

In $p(\bm{\omega} | \mathbf{x}, \mathcal{S})$, both $\mathbf{x}$ and $\mathcal{S}$ are inputs of the prior network. In order to effectively integrate the two conditions, we adopt the cross attention \cite{kim2019attentive} between $\mathbf{x}$ and each element in $\mathcal{S}$. In our case, we have the key-value matrices $K = V \in \mathbb{R}^{C \times d} $, where $d$ is the dimension of the feature representation, and $C$ is the number of categories in the support set. We adopt the instance pooling by taking the average of samples in each category when the shot number $k>1$.

For the query $Q_i = \mathbf{x} \in \mathbb{R}^{d} $, the Laplace kernel returns attentive representation for $\mathbf{x}$:
\begin{equation}
\begin{aligned}
\textbf{Laplace}&(Q_i,K,V) := W_iV \in \mathbb{R}^{d}, \quad W_i := \softmax(-\left\|Q_i-K_{j.} \right\|_{1})^C_{j=1}
\label{Laplace-att}
\end{aligned}
\end{equation}
The prior network takes the attentive representation as the input.

\subsection{More experimental details}
\label{med}
We train all models using the Adam optimizer~\cite{kingma2014adam} with a learning rate of $0.0001$.\ The other training setting and network architecture for regression and classification on three datasets are different as follows.

\subsection{Inference networks}
The architecture of the inference network with vanilla LSTM for the regression task is in Table \ref{inference-regression-1}. The architecture of the inference network with  bidirectional LSTM for the regression task is in Table \ref{inference-regression-2}. For few-shot classification tasks, all models share the same architecture with vanilla LSTM, as in Table \ref{inference-network-1}, For few-shot classification tasks, all models share the same architecture with  bidirectional LSTM, as in Table \ref{inference-network-2}.

\subsection{Prior networks}
The architecture of the prior network for the regression task is in Table \ref{prior-regression}. For few-shot classification tasks, all models share the same architecture, as in Table \ref{prior-network}.

\subsection{Feature embedding networks}

\textbf{Regression.}
The fully connected architecture for regression tasks is shown in Table \ref{fcn-reg}. We train all three models ($3$-shot, $5$-shot, $10$-shot) over a total of $20,000$ iterations, with $6$ episodes per iteration.

\textbf{Classification.}
The CNN architectures for Omniglot, \cifarfs{}, and \mini{} are shown in Table \ref{cnn-Omn}, \ref{cnn-cifar}, and \ref{cnn-mini}. The difference of feature embedding architectures for different datasets is due the different image sizes.

\begin{table}[h]
\small
\begin{center}
    \caption{The inference network $\phi(\cdot)$ based on the vanilla LSTM used for regression.}
	\centering
	\begin{tabular}{cl}
      \toprule
    	\textbf{Output size} & \textbf{Layers} \\
        \midrule
		$40$ & Input samples feature \\
		$40$ & fully connected, ELU \\
		$40$ & fully connected, ELU \\
		$40$ &  LSTM cell, Tanh to $\mu_w$, $\log\sigma^2_w$\\
	 \bottomrule
	\end{tabular}
	\label{inference-regression-1}
	\end{center}
\end{table}

\begin{table}[h]
\small
\begin{center}
    \caption{The inference network $\phi(\cdot)$ based on the bidirectional LSTM for regression. }
	\centering
	\begin{tabular}{cl}
      \toprule
    	\textbf{Output size} & \textbf{Layers} \\
        \midrule
		$80$ & Input samples feature \\
		$40$ & fully connected, ELU \\
		$40$ & fully connected, ELU \\
		$40$ &  LSTM cell, Tanh to $\mu_w$, $\log\sigma^2_w$\\
	
      \bottomrule
	\end{tabular}
	\label{inference-regression-2}
	\end{center}
\end{table}

\begin{table}[h]
\small
\begin{center}
    \caption{The inference network $\phi(\cdot)$ based on the vanilla LSTM for  Omniglot, \mini{}, \cifarfs{}.}
	\centering
	\begin{tabular}{cl}
      \toprule
    	\textbf{Output size} & \textbf{Layers} \\
        \midrule
		$k \times 256$ & Input feature \\
		$256$ & instance pooling \\
		$256$ & fully connected, ELU \\
		$256$ & fully connected, ELU \\
		$256$ &  fully connected, ELU\\
		$256$ &  LSTM cell, tanh to $\mu_w$, $\log\sigma^2_w$ \\
	
      \bottomrule
	\end{tabular}
	\label{inference-network-1}
	\end{center}
\end{table}

\begin{table}[h]
\small
\begin{center}
    \caption{The inference network $\phi(\cdot)$ based on the bidirectional LSTM for  Omniglot, \mini{}, \cifarfs{}.}
	\centering
	\begin{tabular}{cl}
      \toprule
    	\textbf{Output size} & \textbf{Layers} \\
        \midrule
		$k \times 512$ & Input feature \\
		$256$ & instance pooling \\
		$256$ & fully connected, ELU \\
		$256$ & fully connected, ELU \\
		$256$ &  fully connected, ELU\\
		$256$ &  LSTM cell, tanh to $\mu_w$, $\log\sigma^2_w$ \\
	
      \bottomrule
	\end{tabular}
	\label{inference-network-2}
	\end{center}
\end{table}

\begin{table}[h]
\small
\begin{center}
    \caption{The prior network for regression.}
	\centering
	\begin{tabular}{cl}
      \toprule
    	\textbf{Output size} & \textbf{Layers} \\
        \midrule
		$40$ & fully connected, ELU \\
		$40$ & fully connected, ELU \\
	$40$ &  fully connected  to $\mu_w$, $\log\sigma^2_w$ \\
	
      \bottomrule
	\end{tabular}
	\label{prior-regression}
	\end{center}
\end{table}

\begin{table}[h]
\small
\begin{center}
    \caption{The prior network for Omniglot, \mini{}, \cifarfs{}}
	\centering
	\begin{tabular}{cl}
      \toprule
    	\textbf{Output size} & \textbf{Layers} \\
        \midrule
		$256$ & Input query feature\\
		$256$ & fully connected, ELU \\
		$256$ & fully connected, ELU \\
		$256$ &  fully connected  to $\mu_w$, $\log\sigma^2_w$ \\
	
      \bottomrule
	\end{tabular}
	\label{prior-network}
	\end{center}
\end{table}

\begin{table}[h]
\small
\begin{center}
    \caption{The fully connected network $\psi(\cdot)$ used for regression.}
	\centering
	\begin{tabular}{cl}
      \toprule
    	\textbf{Output size} & \textbf{Layers} \\
        \midrule
		$1$ & Input training samples \\
		$40$ & fully connected, RELU \\
		$40$ & fully connected, RELU \\
      \bottomrule
	\end{tabular}
	\label{fcn-reg}
	\end{center}
\end{table}

\begin{table*}[h]
\small
\begin{center}
	\caption{The CNN architecture $\psi(\cdot)$ for Omniglot.}
	\begin{tabular}{p{1.7cm}  p{11.5cm} }
		\hline
		Output size    & Layers \\
		\hline
		$ 28 \mathord\times 28 \mathord\times 1$  &Input images\\\hline
		$ 14 \mathord\times 14 \mathord\times 64$  & \textit{conv2d} ($3 \mathord\times 3$, stride=1, SAME, RELU), dropout 0.9, \textit{pool} ($2 \mathord\times 2$, stride=2, SAME)\\ 
		$ 7 \mathord\times 7 \mathord\times 64$  & \textit{conv2d} ($3 \mathord\times 3$, stride=1, SAME, RELU), dropout 0.9, \textit{pool} ($2 \mathord\times 2$, stride=2, SAME)\\ 
		$ 4 \mathord\times 4 \mathord\times 64$  & \textit{conv2d} ($3 \mathord\times 3$, stride=1, SAME, RELU), dropout 0.9, \textit{pool} ($2 \mathord\times 2$, stride=2, SAME)\\ 
		$ 2 \mathord\times 2 \mathord\times 64$  & \textit{conv2d} ($3 \mathord\times 3$, stride=1, SAME, RELU), dropout 0.9, \textit{pool} ($2 \mathord\times 2$, stride=2, SAME)\\ 
		256 & flatten\\
		\hline
	\end{tabular}
	\label{cnn-Omn}
	\end{center}
\end{table*}

\begin{table*}[h]
\small
\begin{center}
	\caption{The CNN architecture $\psi(\cdot)$  for \cifarfs{}}
	\begin{tabular}{p{1.7cm}  p{11.5cm} }
		\hline
		Output size    & Layers \\
		\hline
		$ 32 \mathord\times 32 \mathord\times 3$  &Input images\\\hline
		$ 16 \mathord\times 16 \mathord\times 64$  & \textit{conv2d} ($3 \mathord\times 3$, stride=1, SAME, RELU), dropout 0.5, \textit{pool} ($2 \mathord\times 2$, stride=2, SAME)\\ 
		$ 8 \mathord\times 8 \mathord\times 64$  & \textit{conv2d} ($3 \mathord\times 3$, stride=1, SAME, RELU), dropout 0.5, \textit{pool} ($2 \mathord\times 2$, stride=2, SAME)\\ 
		$ 4 \mathord\times 4 \mathord\times 64$  & \textit{conv2d} ($3 \mathord\times 3$, stride=1, SAME, RELU), dropout 0.5, \textit{pool} ($2 \mathord\times 2$, stride=2, SAME)\\ 
		$ 2 \mathord\times 2 \mathord\times 64$  & \textit{conv2d} ($3 \mathord\times 3$, stride=1, SAME, RELU), dropout 0.5, \textit{pool} ($2 \mathord\times 2$, stride=2, SAME)\\ 
		256 & flatten\\
		\hline
	\end{tabular}
	\label{cnn-cifar}
	\end{center}
\end{table*}

\begin{table*}[h]
\small
\begin{center}
		\caption{The CNN architecture $\psi(\cdot)$ for \mini{}}
			\begin{tabular}{p{1.7cm}  p{11.5cm} }
				\hline
				Output size    & Layers \\
				\hline
				$ 84 \mathord\times 84 \mathord\times 3$  &Input images\\\hline
				$ 42 \mathord\times 42 \mathord\times 64$  & \textit{conv2d} ($3 \mathord\times 3$, stride=1, SAME, RELU), dropout 0.5, \textit{pool} ($2 \mathord\times 2$, stride=2, SAME)\\ 
				$ 21 \mathord\times 21 \mathord\times 64$  & \textit{conv2d} ($3 \mathord\times 3$, stride=1, SAME, RELU), dropout 0.5, \textit{pool} ($2 \mathord\times 2$, stride=2, SAME)\\ 
				$ 10 \mathord\times 10 \mathord\times 64$  & \textit{conv2d} ($3 \mathord\times 3$, stride=1, SAME, RELU), dropout 0.5, \textit{pool} ($2 \mathord\times 2$, stride=2, SAME)\\ 
				$ 5 \mathord\times 5 \mathord\times 64$  & \textit{conv2d} ($3 \mathord\times 3$, stride=1, SAME, RELU), dropout 0.5, \textit{pool} ($2 \mathord\times 2$, stride=2, SAME)\\ 
				$ 2 \mathord\times 2 \mathord\times 64$  & \textit{conv2d} ($3 \mathord\times 3$, stride=1, SAME, RELU), dropout 0.5, \textit{pool} ($2 \mathord\times 2$, stride=2, SAME)\\ 
				256 & flatten\\
				\hline
			\end{tabular}
			\label{cnn-mini}
			\end{center}
\end{table*}

\subsection{Other settings}
The settings including the iteration numbers and the batch sizes are different on different datasets. The detailed information is given in Table \ref{opt}.

\begin{table}[h]
\small
\begin{center}
    \caption{The iteration numbers and batch sizes on different datasets.}
	\centering
	\begin{tabular}{lrr}
      \toprule
    	\textbf{Dataset} & \textbf{Iteration} & \textbf{Batch size} \\
        \midrule
		Regression & $20,000$ & $25$ \\
		 Omniglot & $100,000$ & $6$ \\
		\cifarfs{} & $200,000$ & $8$ \\
		\mini &  $150,000$ & $8$ \\
      \bottomrule
	\end{tabular}
	\label{opt}
	\end{center}
\end{table}

\end{document}